\documentclass[twocolumn]{autart}
\input epsf
\usepackage{cite,epsfig}
\usepackage{graphicx,amsmath,amssymb,remark,color}
\usepackage[T1]{fontenc}
\usepackage[latin1]{inputenc}
\usepackage{graphicx}
\usepackage{amssymb}
\usepackage{amsmath}
\usepackage{dcolumn}
\usepackage{bm}
\usepackage{color}
\usepackage{multicol}
\usepackage{algorithm}
\usepackage{multirow}
\usepackage{subfigure} 
\usepackage{hyperref}
\usepackage{xcolor}
\usepackage{xpatch}
\usepackage{algorithm}
\usepackage{algpseudocode} 
\input epsf
\usepackage{cite,epsfig}
\usepackage{graphicx,amsmath,amssymb,remark,color}
\newcommand{\EQ}{\begin{eqnarray}}
\newcommand{\EN}{\end{eqnarray}}
\newcommand{\EQQ}{\begin{eqnarray*}}
\newcommand{\ENN}{\end{eqnarray*}}
\def\widebar{\accentset{{\cc@style\underline{\mskip10mu}}}}

 \makeatletter
\newcommand*\bigcdot{\mathpalette\bigcdot@{.5}}
\newcommand*\bigcdot@[2]{\mathbin{\vcenter{\hbox{\scalebox{#2}{$\m@th#1\bullet$}}}}}
\makeatother

\DeclareMathAlphabet{\matheur}{U}{eur}{m}{n}
\DeclareMathAlphabet{\matheurb}{U}{eur}{b}{n}
\DeclareMathAlphabet{\matheus}{U}{eus}{m}{n}
\DeclareMathAlphabet{\matheuf}{U}{euf}{m}{n}
\renewcommand{\t}{^{\mbox{\tiny\sf T}}}

\newcommand{\eproof}{\hfill\rule{2mm}{2mm}}

\newcommand{\bremark}{\begin{remark}
\begin{rm}}
\newcommand{\eremark}{ \end{rm}
\end{remark} }
\newcommand{\btheorem}{\begin{theorem} \begin{it}}
\newcommand{\etheorem}{\end{it} \hfill \rule{1mm}{2mm}
\end{theorem} }
\newcommand{\blemma}{\begin{lemma} \begin{it} }
\newcommand{\elemma}{ \end{it} \hfill\rule{1mm}{2mm}
\end{lemma} }
\newcommand{\bcorollary}{\begin{corollary} \begin{it} }
\newcommand{\ecorollary}{ \end{it} \hfill\rule{1mm}{2mm}
\end{corollary} }
\newcommand{\bdefinition}{\begin{definition} }
\newcommand{\edefinition}{ \hfill\rule{1mm}{2mm}
\end{definition} }
\newcommand{\bproposition}{\begin{proposition} }
\newcommand{\eproposition}{\hfill \rule{1mm}{2mm}
\end{proposition} }
\newcommand{\bexample}{\begin{example} \begin{rm}}
\newcommand{\eexample}{ \end{rm} \hfill\rule{1mm}{2mm}
\end{example} }

\newcommand{\basm}{\begin{assumption} \begin{rm} }
\newcommand{\easm}{ \end{rm} \hfill\rule{1mm}{2mm}
\end{assumption} }

\begin{document}

\newtheorem{theorem}{\sf\bfseries Theorem}[section]
\newtheorem{lemma}{\sf\bfseries Lemma}[section]
\newtheorem{coro}{\sf\bfseries Corollary}[section]
\newtheorem{definition}{\sf\bfseries Definition}[section]
\newtheorem{remark}{\sf\bfseries Remark}[section]
\newtheorem{corollary}{\sf\bfseries Corollary}[section]
\newtheorem{proposition}{\sf\bfseries Proposition}[section]
\newtheorem{example}{\sf\bfseries Example}[section]
\newtheorem{assumption}{\sf\bfseries Assumption}
\newtheorem{problem}{\sf\bfseries Problem}

\begin{frontmatter}

\title{C2TE: Coordinated Constrained Task Execution Design for Ordering-Flexible Multi-Vehicle Platoon Merging}

\thanks{
This work was
supported in part by A*STAR AME Young Individual Research under
Grant A2084c0156, and in part by the SUG-NAP Grant of Nanyang
Technological University. 
Email addresses:  {\tt \{binbin.hu, yanxin.zhou, shuo.cheng, lyuchen\}@ntu.edu.sg, weihenglai@gmail.com}.\\
$^*$~Corresponding author.
}

\author[ntu]{Bin-Bin Hu},
\author[ntu]{Yanxin Zhou}, 
\author[bh]{Henglai Wei},
\author[ntu]{Shuo Cheng}, 
\author[ntu]{Chen Lv}$^*$,


\address[ntu]
{School of Mechanical and Aerospace Engineering,Nanyang Technological University, Singapore, 637460}
\address[bh]
{School of Transportation Science and Engineering, Beihang University, Beijing, China, 100191}

\begin{abstract}                          
In this paper, we propose a distributed coordinated constrained task execution (C2TE) algorithm that enables a team of vehicles from different lanes to cooperatively merge into an {\it ordering-flexible platoon} maneuvering on the desired lane. Therein, the platoon is flexible in the sense that no specific spatial ordering sequences of vehicles are predetermined. To attain such a flexible platoon, we first separate the multi-vehicle platoon (MVP) merging mission into two stages, namely, pre-merging regulation and {\it ordering-flexible platoon} merging, and then formulate them into distributed constraint-based optimization problems. Particularly, by encoding longitudinal-distance regulation and same-lane collision avoidance subtasks into the corresponding control barrier function (CBF) constraints, the proposed algorithm in Stage 1 can safely enlarge sufficient longitudinal distances among adjacent vehicles. Then, by encoding lateral convergence, longitudinal-target attraction, and neighboring collision avoidance subtasks into CBF constraints, the proposed algorithm in Stage~2 can efficiently achieve the {\it ordering-flexible platoon}. Note that the {\it ordering-flexible platoon} is realized through the interaction of the longitudinal-target attraction and time-varying neighboring collision avoidance constraints simultaneously. Feasibility guarantee and rigorous convergence analysis are both provided under strong nonlinear couplings induced by flexible orderings. Finally, experiments using three autonomous mobile vehicles (AMVs) are conducted to verify the effectiveness and flexibility of the proposed algorithm, and extensive simulations are performed to demonstrate its robustness, adaptability, and scalability when tackling vehicles' sudden breakdown, new appearing, different number of lanes, mixed autonomy, and large-scale scenarios, respectively.

\end{abstract}

\end{frontmatter}

\section{Introduction}
In recent years, it has witnessed a significant development of multi-vehicle coordination, including connected driving, coordinated navigation, collective convoying, etc~\cite{jin2020analysis,hu2023coordinated,hu2023cooperative}. Therein, due to the benefits of improved efficiency,  promoted capacity, and enhanced stability, the research of multi-vehicle platoon (MVP) merging has also received considerable attention, which is essentially to govern vehicles to cooperatively merge into a platoon on the desired lane~\cite{lu2003longitudinal}. 
According to whether the platoon is achieved with a fixed spatial ordering or not, the existing MVP merging works can be approximately separated into two categories, namely, ordering-fixed and {\it ordering-flexible} merging approaches.

For the ordering-fixed merging approaches, the core idea is to govern the vehicles to merge into a predetermined gap such that the vehicles in the platoon keep a fixed ordering sequence \cite{milanes2010automated}, which is straightforward and convenient to implement in practice.
As pioneer works, the initial efforts have been devoted to a dynamic adaptive merging algorithm for on-road vehicles \cite{karbalaieali2019dynamic}. Later, it was extended to automated industrial vehicles \cite{liu2022cooperative}. However, these two works \cite{karbalaieali2019dynamic, liu2022cooperative} only considered simple one-vehicle merging, which does not touch the more sophisticated scenario of vehicle joining and leaving. To address this problem, a distributed MPC strategy was proposed by \cite{liu2018distributed,kazemi2024longitudinal} for one-vehicle joining and leaving cases. A velocity-obstacle controller was designed in \cite{wang2023collision} to address the additional collision avoidance. A signal-temporal-logic controller was developed in \cite{charitidou2022splitting}. 
For more complex multiple-vehicle merging, some optimization methods were developed in \cite{wei2023hierarchical}.  For some special applications, considerable efforts have been devoted to the MPV merging on high on-ramp situations \cite{kumaravel2021decentralized}. In this effort, a control strategy was proposed in \cite{scholte2022control} for one-vehicle merging on the ramp. Later, an optimal control using the control barrier function (CBF) was proposed in \cite{xu2022feasibility} to provide safety guarantees. However, the aforementioned works \cite{milanes2010automated,karbalaieali2019dynamic,liu2022cooperative,liu2018distributed,wang2023collision,charitidou2022splitting,kazemi2024longitudinal,wei2023hierarchical,kumaravel2021decentralized,scholte2022control,xu2022feasibility} predetermine the vehicle orderings with fixed neighborhoods, which render the platoon lack flexibility and may result in longer travel distances.

For the {\it ordering-flexible} methods in achieving MVP merging mission, the vehicles can efficiently merge into arbitrary gaps on the desired lane according to proximity. Note that, the {\it ordering-flexible platoon} relies on the time-varying vehicle interaction, which renders the platoon with flexible orderings but may pose challenging issues to the convergence of MVP merging missions. In this pursuit, a distributed hybrid control law was developed in \cite{lan2011synthesis} to coordinate the vehicles to form a platoon moving along the desired paths with local convergence. Later, a distributed guiding-vector-field algorithm was designed in \cite{hu2023spontaneous} to form an {\it ordering-flexible platoon} with global convergence. Despite that the aforementioned works \cite{lan2011synthesis,hu2023spontaneous} have tried to address {\it ordering-flexible} platoon problem, they have not considered the feasibility guarantee of such a problem due to the sophisticated nonlinear couplings induced by flexible orderings.  Moreover, the previous works \cite{lan2011synthesis,hu2023spontaneous} only considered the simple robot dynamics in unstructured static environments, once the structure lanes and changing elements including vehicles sudden breakdown and newly appearing are added to environments, the {\it odering-flexible} MVP merging problem will become more intricate. Therefore, the feasibility guarantee and performance of {\it ordering-flexible} MVP merging in structure lanes remain an open question.

To this end, inspired by the constraint-based task execution framework \cite{notomista2021resilient} and {\it ordering-flexible} mechanism~\cite{hu2023spontaneous}, we design a distributed C2TE algorithm that enables a team of vehicles from different lanes to cooperatively merge into a platoon of an arbitrary ordering sequence maneuvering on the desired lane. 
In particular, by encoding the longitudinal-distance regulation and same-lane collision avoidance subtasks into CBF constraints, the proposed algorithm in Stage 1 can safely enlarge sufficient longitudinal distances among adjacent vehicles. Then, by encoding lateral convergence, longitudinal-target attraction, and neighboring collision avoidance subtasks into CBF constraints, the proposed algorithm in Stage~2 can efficiently achieve the {\it ordering-flexible platoon}. Therein, the interaction between longitudinal-target-convergence constraints and time-varying neighboring collision-avoidance constraints further results in the platoon of arbitrary orderings. Finally, experiments and simulations are conducted to verify the effectiveness, flexibility, robustness, adaptability, and scalability of the proposed algorithm. The main contribution of this paper is twofold.

\begin{enumerate}
 \item We propose a distributed C2TE algorithm that enables a team of vehicles from different lanes to cooperatively enlarge sufficient longitudinal distances, and then efficiently and safely merge into a platoon maneuvering on the desired structural lane with an arbitrary ordering sequence. 
 
 \item We rigorously guarantee the asymptotical convergence of the {\it ordering-flexible} platoon merging under strong nonlinear couplings induced by interaction of longitudinal-target attraction and time-varying neighboring collision avoidance constraints, simultaneously.

\end{enumerate}

The remainder of this paper is organized below. Section~2 introduces some preliminaries. Section 3 formulates the {\it ordering-flexible} MVP merging problem. Section 4 illustrates the C2TE optimization design. Section~5 elaborates on the feasibility guarantee and convergence analysis.
Section 6 provides experiments and extensive simulations. Finally, the conclusion is drawn in Section 7.

Throughout the paper, the real and positive real numbers are denoted by $\mathbb{R},\mathbb{R}^+$, respectively. The $n$-dimensional Euclidean space is denoted by $\mathbb{R}^n$. The integer numbers are denoted by $\mathbb{Z}$. The notation $\mathbb{Z}_i^j$ represents the integer set $\{m\in \mathbb{Z}~|~i\leq m\leq j\}$. The $n$-dimensional column vector of all 0's is denoted by $\mathbf{0}_n$.

\section{Preliminaries}

\subsection{Target Set}
Suppose a desired task set $\mathcal T \subseteq\mathbb{R}^n$ is defined by 
\begin{align}
\label{target_set}
\mathcal T:=\{\xi\in\mathbb{R}^n~|~\phi(\xi)\leq0\},
\end{align}
where $\xi:=[\xi_1, \dots, \xi_n]\t$ are the dummy coordinates in the $n$-dimensional Euclidean space and  $\phi: \mathbb{R}^{n}\rightarrow\mathbb{R}$ is a twice continuously differentiable function. By substituting a state $p_0$ for $\xi$ in \eqref{target_set}, the function $\phi(p_0)$ is to illustrate the task-convergence errors between $\mathcal{T}$ and $p_0$. Essentially, $-\phi(\xi)$ can be regarded as a CBF function and will be used to establish CBF constraints \cite{ames2016control} later.


\subsection{Desired Lane}
The first building block is the desired lane $\mathcal L^d$. Analogous to 
$\mathcal T$ in \eqref{target_set}, $\mathcal L^d$ is characterized by
\begin{align}
\label{desired_lane}
\mathcal L^d :=\{ \xi\in\mathbb{R}~|~\phi(\xi):=|\xi-y_d|=0 \},
\end{align}
where the constant $y_d\in\mathbb{R}$ is tracked by the lateral position of each vehicle. The interpretation of  $\mathcal L^d$ in \eqref{desired_lane} will be utilized to establish CBF constraint \eqref{stage_2_lateral_converge_CBF} later.

\subsection{Multi-Vehicle Systems}
The second building block is a team of vehicles $\mathcal V=\{1,2, \cdots, N\}$, of which each vehicle is moving according to the bike dynamics in the Cartesian coordinates~\cite{dieter2018vehicle},
\begin{figure}[!htb]
  \centering
  \includegraphics[width=5.0cm]{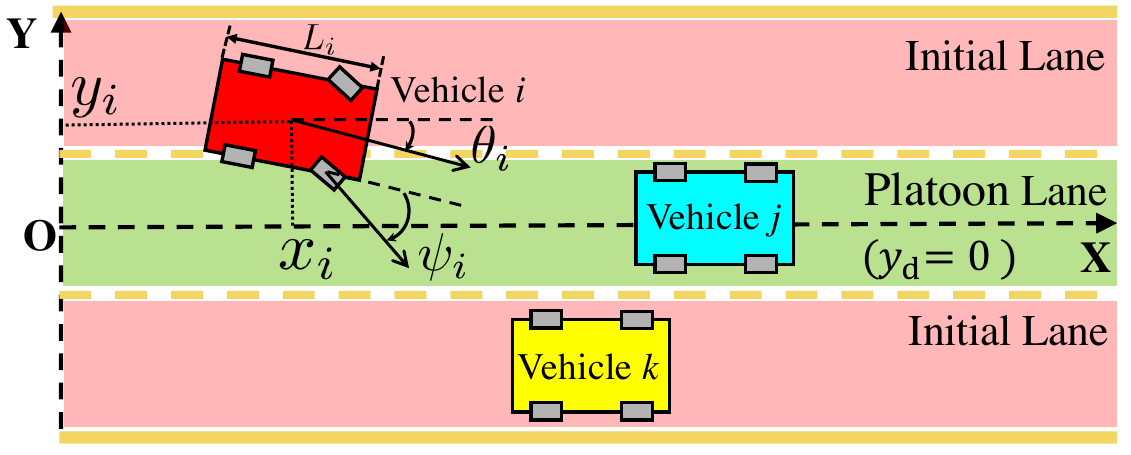}
  \caption{Illustration of vehicle kinematics and different lanes. The pink and green lanes are the initial and platoon merging lanes, respectively.}
  \label{Vehicle_Lane}
\end{figure} 
\begin{align}
\label{vehicle_dynamics}
\dot{x}_i=&v_i\cos\theta_i,~\dot{y}_i=v_i\sin\theta_i,~\dot{\theta}_i= v_i\frac{\tan \psi_i}{ B_i}, i\in\mathcal V,
\end{align} 
where $p_i=[x_i, y_i]\t\in\mathbb{R}^2, v_i\in\mathbb{R}, \theta_i\in[-\pi, \pi)$ are the positions, body velocity, and yaw angle of vehicle $i$ in 2D, respectively, and $\psi_i, B_i\in\mathbb{R}^+$ denote the steering angle of front wheels and the length of vehicle base, respectively in Fig.~\ref{Vehicle_Lane}.

\subsection{Constraint-Based Task Execution  }
The third building block is the constraint-based task execution, which can be utilized to encode the MVP merging into an online optimization with different subtasks.

\begin{definition}
\label{definition_constraint_based}
(Constraint-based task execution) \cite{notomista2021resilient}. For a desired task set $\mathcal T$ in \eqref{target_set} and a vehicle governed by the single-integrator dynamic $\dot{x}=u$
with the states $x\in\mathbb{R}^n$, and the inputs $u\in\mathbb{R}^n$,  the constraint-based task execution can be achieved by an arbitrary Lipschitz continuous controller $u^{\ast}$ from the following constrained optimization problem 
\begin{align}
\label{constraint_minimization}
\min_{u, \delta}~\{\|u\|^2+|\delta|^2\}~\mathrm{s.t.}~&\frac{\partial h(x)}{\partial x}u+\gamma(x)-\delta\leq 0, 
\end{align}
which features the asymptotical convergence and forward invariance, i.e.,
\begin{align}
\label{property}
&x(0)\notin\mathcal T \implies x(t)\rightarrow\in\mathcal T~\mathrm{as}~t\rightarrow\infty,\nonumber\\
&x(T)\in\mathcal T(\mathrm{i.e.}, \delta=0) \implies x(t)\in\mathcal T, \forall t>T.  
\end{align}
Here, (i) the CBF function $h(x)$ adheres to $h(x) = -\psi(x) \geq 0$, where $\psi(x) \geq 0$ represents a cost function \cite{notomista2021resilient}. Furthermore, the CBF functions incorporating additional slack variables in \eqref{stage_1_longitudinal_regulation_CBF}, \eqref{stage_2_lateral_converge_CBF}, \eqref{stage_2_longitudinal_merging_CBF} also comply with such a condition later. It has been shown in Definition 2 of \cite{ames2016control} that the continuous function $\gamma(\cdot):(-b, a)\rightarrow(-\infty, +\infty)$ is an extended class $\mathcal K$ function for $a,b>0$, and strictly increasing and satisfies $\gamma(0)=0$, which slightly differs from the traditional class $\mathcal K$ function \cite{khalil2002nonlinear}.~(ii)~For the forward invariance in \eqref{property}, it only holds if $\delta=0$. When $x(T)\in\mathcal T$ at some time $T$, it follows from the definition of CBF that  $h(x)=-\psi(x)\geq0$, i.e., $\psi(x)\leq0$. Meanwhile, recalling $\psi(x)\geq0$ in (i), one has that 
$x\in\mathcal T\Rightarrow h(x)=\psi(x)=0,$
which implies that $\gamma(h(x))=0$ as well. By substituting $h(x)=0$ and $\gamma(h(x))=0$ into the constraint in (4) yields $-\delta\leq0$, which follows from the minimization of $|\delta|^2$ in \eqref{constraint_minimization} that the optimal $\delta^{\ast}=0$. Accordingly, when $\delta=0$, the constraint-based task execution in \eqref{constraint_minimization} will degenerate into the traditional CBF-quadratic-program (CBF-QP), which thus features forward invariance \cite{ames2016control}.
\end{definition}


In Definition~\ref{definition_constraint_based}, the forward invariance in \eqref{property} can be utilized to guarantee collision avoidance all along, which avoids the sophisticated analysis of traditional string stability~\cite{han2023prescribed}. Moreover, the constraint-based task execution in \eqref{constraint_minimization} always contains a feasible solution $\{u=\mathbf{0}_n, \delta~\mathrm{~is~sufficiently~large}\}$, which are utilized to guarantee the {\it ordering-flexible platoon} later.

 \begin{remark}
\label{remark_boundary}
Using the Karush-Kuhn-Tucker (KKT) conditions \cite{boyd2004convex} for \eqref{constraint_minimization}, it has been shown in \cite{notomista2019constraint} that
the optimal solutions $u^{\ast}, \delta^{\ast}$ and the optimal Lagrange multiplier $\lambda^{\ast}\geq0$ associated with the constraint  $g:=-({\partial h(x)}/{\partial x\t})u-\gamma\big(h(x)\big)+\delta$ in~(\ref{constraint_minimization}) are $ u^{\ast}$  $ =-\frac{\partial h(x)}{\partial x}{\gamma(h(x))}/{(1+\|\frac{\partial h(x)}{\partial x\t}\|^2)},~\lambda^{\ast}={-2\gamma(h(x))}/(1$ $+\|\frac{\partial h(x)}{\partial x\t}\|^2), \delta^{\ast}={-\gamma(h(x))}/{(1+\|\frac{\partial h(x)}{\partial x\t}\|^2)}$.
Due to the properties of the extended class $\mathcal K$ function $\gamma(\cdot)$ in~\eqref{constraint_minimization}, one has that 1) $\lambda^{\ast}>0\iff h(x)<0 \iff x \notin \mathcal{T}$, and 2) $\lambda^{\ast}=0 \iff  h(x)=0 \implies x \in \mathcal{T}$.
From $\lambda^{\ast}g=0$ in the complementary slackness condition, the boundary of constraint in (\ref{constraint_minimization}) is activated if $\lambda^{\ast}>0$, i.e., $\lambda^{\ast}>0 \Rightarrow g=0$. Meanwhile, since $h(x)=-\phi(x)$ and $\mathcal T:=\{\xi\in\mathbb{R}^n~\big|~\phi(\xi)\leq 0\}$ are given in~\eqref{target_set}, one has that 
\begin{align}
\label{proprtty1}
&\lambda^{\ast}>0 \iff x(t)\notin \mathcal T \implies g=0,\nonumber\\
&\lambda^{\ast}=0\implies x(t)\in \mathcal T\implies g\leq0.
\end{align}
Such relationship of~~$\lambda^{\ast}$ and $x(t) \in\mathcal T$ in~\eqref{proprtty1} will be utilized in the proof of Lemma~\ref{stage_2_lemma_longitudinal_coolision} later.
 \end{remark}

\section{Problem Formulation}
\label{sec_problem_formulation}

\subsection{Pre-Merging Regulation}
\label{sub_pre_merging}
For some vehicles initially on different lanes, it is crucial for these vehicles to enlarge sufficient longitudinal distances before merging into the desired lane $\mathcal L_d$, which motivates the pre-merging regulation in Stage 1.

\begin{definition}
\label{def_pre_merging}
(Pre-merging regulation)
All vehicles $\mathcal V$ governed by \eqref{vehicle_dynamics} collectively achieves the pre-merging regulation if the following three objectives are fulfilled,

\begin{enumerate}
\item \label{Pre_platoon_obj_1} {\sf\bfseries(Longitudinal-distance regulation)}: The longitudinal distance between vehicle $i$ and other vehicles asymptotically converges to be greater than the sensing distance $R$, i.e., $\lim_{t\rightarrow\infty}|x_{i}(t)-x_j(t)|\geq R, \forall j\neq i\in\mathcal V,$
where $R>B_i$ with $B_i$ being the length of vehicle~$i$ in~\eqref{vehicle_dynamics}.

\item \label{Pre_platoon_obj_2} {\sf\bfseries (Lateral-lane keeping)}: Each vehicle $i$ keeps its lateral lane until it achieves sufficient longitudinal spacing, i.e.,
$\exists T_1>0~\mathrm{satisfying}~|x_{i}(T_1)-x_j(T_1)|> \rho,   j\neq i\in\mathcal V,~\mathrm{such~that}~y_i(t)=y_i(0), \forall t\in(0, T_1),$
where $y_i(t)$ and $y_i(0)$ denote the lateral positions of vehicle $i$ at time $t > 0$ and $t = 0$, respectively, and $\rho \in (r, R)$ is a positive constant with $R$ and $r$ being the sensing and safe radii, respectively.

\item \label{Pre_platoon_obj_3} {\sf\bfseries (Same-lane collision avoidance)}: The collision avoidance between arbitrary two vehicles $i,l$ on the same lane is guaranteed all along, i.e., $|x_{i}(t)-x_l(t)|\geq r, \forall t\geq0, \{ i\neq l~|~y_i=y_l\},$
where $r\in(B_i,  R), \forall i\in\mathcal V$ is the safe distance with $B_i$ being the length of vehicle~$i$ in~\eqref{vehicle_dynamics}.

\begin{figure}[!htb]
  \centering
  \includegraphics[width=6cm]{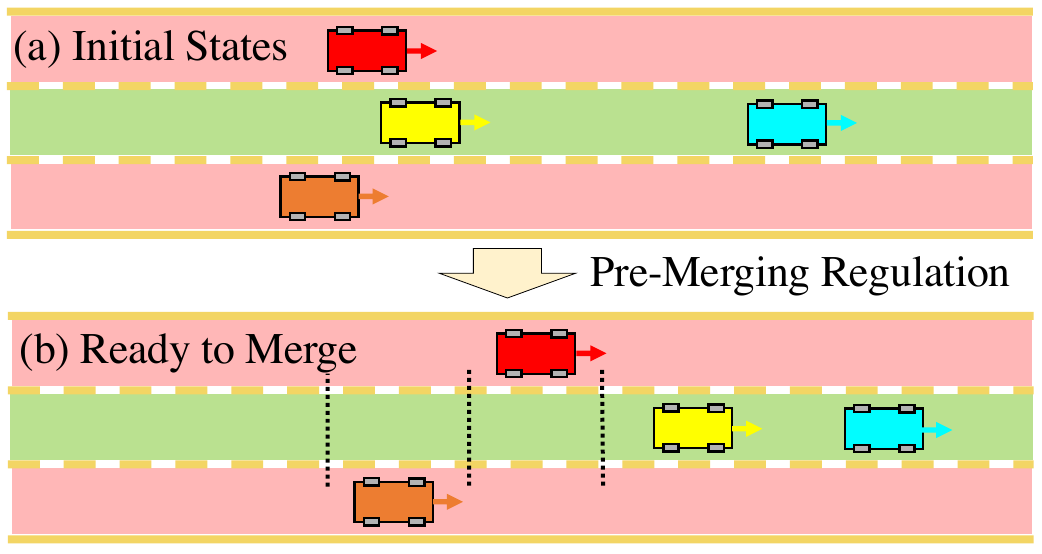}
  \caption{Illustration of the pre-merging regulation in Definition~\ref{def_pre_merging} to enlarge sufficient longitudinal distances.}
  \label{stage_1_illustration}
\end{figure}

\end{enumerate}
\end{definition}
In Definition~\ref{def_pre_merging}, Objective~\ref{Pre_platoon_obj_2}) ensures a smooth transition to the subsequent lane-merging control once there exists sufficient longitudinal distances. From Objective~\ref{Pre_platoon_obj_2}), the collision avoidance only needs to be guaranteed in the same lane in Objective \ref{Pre_platoon_obj_3}), as shown in Fig.~\ref{stage_1_illustration}.


\subsection{Ordering-Flexible MVP Merging}
\label{sub_multi_vehicle_merging}
Once each vehicle $i$ gets sufficient longitudinal distances in Section~\ref{sub_pre_merging}, we can conduct the multi-vehicle lateral merging missions.
Analogous to~\eqref{desired_lane}, suppose that the common lane $\mathcal L_i^d$ for vehicle $i$ is 
\begin{align}
\label{vehicle_i_desired_lane}
\mathcal L_i^d :=\{ \xi_i\in\mathbb{R}~|~\phi_i(\xi_i):=|\xi_i-y_d|=0 \}, \forall i \in\mathcal V,
\end{align}
with $\xi_i$ and $\phi_i(\xi_i)$ being the dummy coordinates and implicit functions, respectively. Note that the target set of $\mathcal L_i^d$ only contains one common point $\{y_d\}$.
Subscribing $y_i$ in \eqref{vehicle_dynamics} for $\xi_i$ in \eqref{vehicle_i_desired_lane}, the lateral-convergence errors between vehicle $i$ and $\mathcal L_i^d$ become $\phi_{i}(y_i)=|y_i-y_d|, \forall i\in\mathcal V$, which implies that the multi-vehicle lateral merging is achieved once if the lateral-convergence errors converge to zeros, i.e., $\lim_{t\rightarrow\infty}\phi_{i}(y_i(t))=0, \forall i\in\mathcal V.$



\begin{definition}
\label{def_platoon}
({\it Ordering-flexible} MVP merging) All vehicles~$\mathcal V$ governed by~\eqref{vehicle_dynamics} and Definition~\ref{def_pre_merging} collectively achieve the {\it ordering-flexible} MVP merging on the desired lane $\mathcal L_i^d$ in~\eqref{vehicle_i_desired_lane}, if the following four objectives are fulfilled,
\begin{enumerate}
\item \label{platoon_obj_1} {\sf\bfseries (Lateral convergence)}: All vehicles asymptotically merge to the desired common lane $\mathcal L_i^d$, i.e., 
$\lim_{t\rightarrow\infty}y_i(t)=y_d, \forall i\in \mathcal V$
with $y_d$ given in~\eqref{desired_lane}.

\item \label{platoon_obj_2} {\sf\bfseries(Ordering-flexible platoon)}: The vehicle platoon is achieved with arbitrary spatial ordering sequences in the longitudinal direction, i.e, $r<\lim_{t\rightarrow\infty}|x_{s[k]}(t)-x_{s[k+1]}(t)|<\rho, k\in\mathbb{Z}_1^{N-1},$
where $\rho, r, R$ are given in Definition~\ref{def_pre_merging}, $s[1], \cdots, s[N]$ are the spatial ordering of the indexes from the head to the tail of the platoon, and $x_{s[k]}$ is the longitudinal position of $s[k]$-th vehicle in \eqref{vehicle_dynamics}.


\item \label{platoon_obj_3} {\sf\bfseries(Platoon cruising)}: All vehicles maintain the relative distances and cruise with a desired longitudinal velocity, i.e., $\lim_{t\rightarrow\infty}\dot{x}_i(t)=\lim_{t\rightarrow\infty}\dot{x}_j(t)=v_d, \forall i\neq j\in\mathcal V$,
where $v_d\in\mathbb{R}^+$ is a desired velocity of a virtual target in \eqref{virtual_target_dynamics} later.
\begin{figure}[!htb]
  \centering
  \includegraphics[width=6cm]{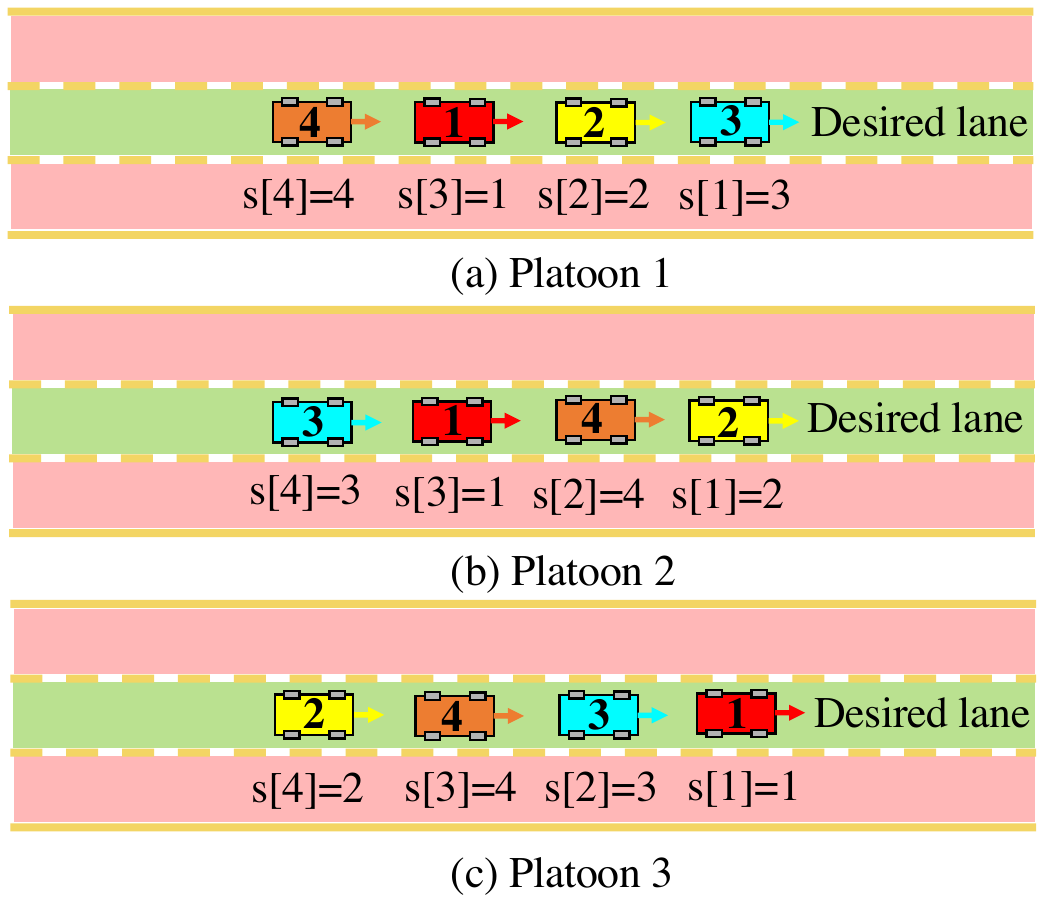}
  \caption{Three examples of the four-vehicle platoon with different ordering sequences in Definition~\ref{def_platoon}.}
  \label{OF_platoon}
\end{figure}

\item \label{platoon_obj_4} {\sf\bfseries (Collision avoidance)} The inter-vehicle collision avoidance is guaranteed all along, i.e., $|x_i(t)-x_j(t)|\geq r, \forall i\neq j\in\mathcal V,  t\geq0$.
 \end{enumerate}
\end{definition}

In Definition~\ref{def_platoon}, Objectives \ref{platoon_obj_1})-\ref{platoon_obj_3}) ensure the platoon with flexible orderings. Examples are given in Fig.~\ref{OF_platoon}.

\subsection{Problem Formulation}
\label{sub_pro_formulation}
 
In this section, since the velocity and steering $v_i$ and $\psi_i$ in \eqref{vehicle_dynamics} fail to control the longitudinal and lateral motion in Definitions~\ref{def_pre_merging} and \ref{def_platoon} directly, we first bridge this gap and then formulate the problem.
Let $p_i^o:=[x_i^o, y_i^o]\in\mathbb{R}^2$ be a virtual point orthogonal to the wheel axis and near the vehicle center~\cite{glotfelter2019hybrid}, i.e., 
\begin{align}
\label{virtual_point}
p_i^o=p_i+d R(\theta_i)e_1
\end{align}
with  $p_i, \theta_i$ given in~\eqref{vehicle_dynamics}, a small constant $d\in\mathbb{R}^+$, and  
\begin{align}
R(\theta_i)=
\begin{bmatrix}
\cos\theta_i & -\sin\theta_i \\
\sin\theta_i & \cos\theta_i \\
\end{bmatrix},
e_1=[1, 0]\t.
\end{align}
Let $u_{i,x}^{r}\in\mathbb{R}, u_{i,y}^{r}\in\mathbb{R}$ be the desired  longitudinal and lateral inputs for the virtual point $p_i^o$ in \eqref{virtual_point}, one has that 
\begin{align}
\label{virtual_derivative}
\dot{x}_i^o=u_{i,x}^r,~\dot{y}_i^o=u_{i,y}^r.
\end{align}
Taking the derivative of \eqref{virtual_point} along the dynamics of \eqref{virtual_derivative},
and substituting $\dot{p}_i, \theta_i$ in \eqref{vehicle_dynamics} into \eqref{virtual_derivative} yields 
\begin{align}
\begin{bmatrix}
u_{i,x}^r\\
u_{i,y}^r
\end{bmatrix}
=&\begin{bmatrix}
\cos\theta_i & -d\sin\theta_i\\
\sin\theta_i & d\cos\theta_i
\end{bmatrix}
\begin{bmatrix}
v_i\\
v_i\frac{\tan \psi_i}{B_i}
\end{bmatrix},
\end{align}
which implies that the actual inputs $v_i$ and $\psi_i$ in \eqref{vehicle_dynamics} are 
\begin{figure}[!htb]
  \centering
  \includegraphics[width=7.9cm]{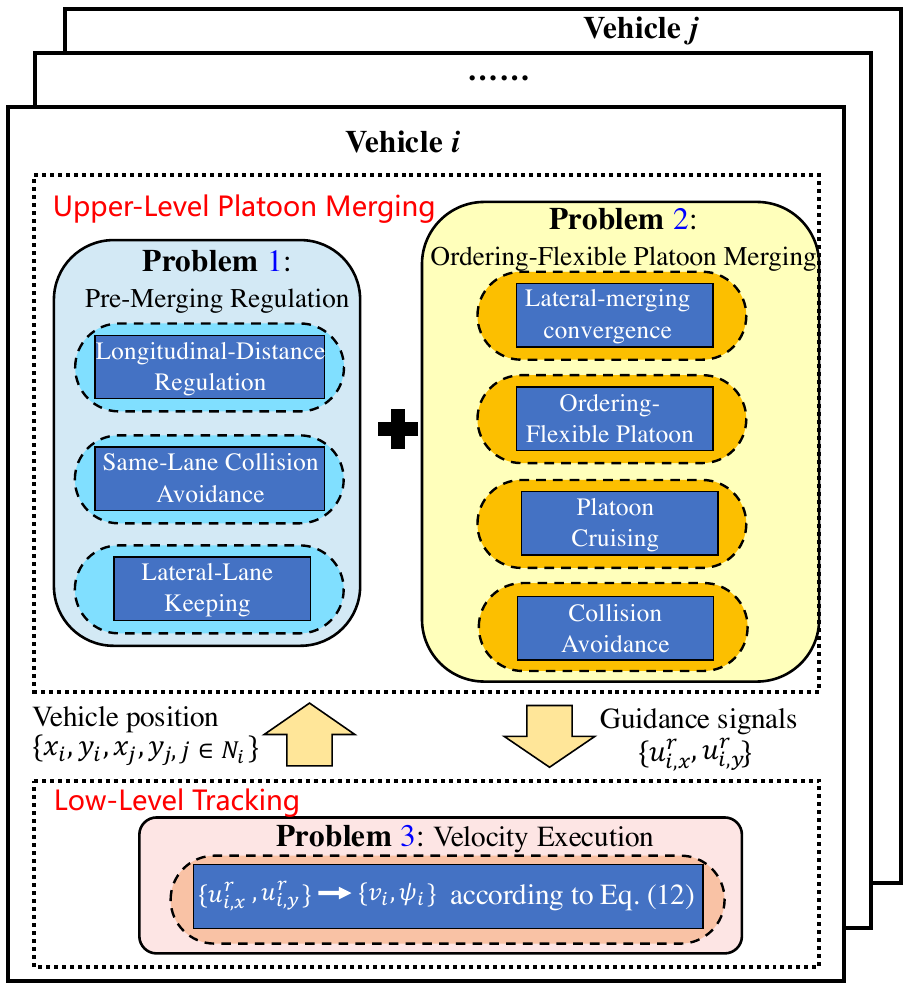}
  \caption{A hierarchical architecture of Problems~\ref{problem_1}-\ref{problem_3}. }
  \label{platoon_struture}
\end{figure}
\begin{align}
\label{velocity_convertion}
v_i=&u_{i,x}^{r} \cos{\theta}_i + u_{i,y}^{r} \sin{\theta}_i,\nonumber\\
\psi_i=&\arctan \frac{(-u_{i,x}^{r}\frac{\sin{\theta_i}}{d}+u_{i,y}^{r}\frac{\cos{\theta_i}}{d}) B_i}{u_{i,x}^{r} \cos{\theta}_i + u_{i,y}^{r} \sin{\theta}_i}
\end{align} with $u_{i,x}^r, u_{i,y}^r$ given in~\eqref{virtual_derivative}. 
The undesired scenario of $u_{i,x}=0, u_{i,y}=0$ is excluded. If $u_{i,x}=0, u_{i,y}=0$, vehicle $i$ actually does not move or change the yaw angle $\psi_i$, there is no need to calculate $v_i, \psi_i$ via \eqref{velocity_convertion}. 
Since $p_i^o$ in \eqref{virtual_point} is near the vehicle center $p_i$ in \eqref{vehicle_dynamics}, we can leverage the near-identity diffeomorphism~\cite{glotfelter2019hybrid} to control vehicle $i$ using the decoupled inputs $u_{i,x}^r$ and $u_{i,y}^r$ in \eqref{virtual_derivative}, i.e., 
\begin{align}
\label{decouple_dynamics}
\dot{x}_i= u_{i,x}^r, \dot{y}_i= u_{i,y}^r.
\end{align} 


\begin{problem}
\label{problem_1}
(Pre-Merging Control) Design the desired velocities $\{u_{i,x}^r, u_{i,y}^r\}, i\in\mathcal V,$
for each vehicle $i, i\in\mathcal V$ governed by \eqref{decouple_dynamics} such that it gets sufficient longitudinal distances, i.e., Objectives \ref{Pre_platoon_obj_1})-\ref{Pre_platoon_obj_3}) in Definition~\ref{def_pre_merging}.
\end{problem}

Since the objective of $\lim_{t\rightarrow\infty}|\  x_{i}-x_{j}|\geq R, \forall i\neq j\in\mathcal V,$ in Definition~\ref{def_pre_merging} has been asymptotically achieved in Problem~\ref{problem_1}, there always exists a positive time $T_1\in\mathbb{R}^+$ such that $\lim_{t\rightarrow T_1}| x_{i}-x_{j}|\geq \rho$ because of $R>\rho$, which implies that vehicle $i$ is ready to perform the {\it ordering-flexible} MVP merging in Problem~2.

\begin{problem}
\label{problem_2}
(Ordering-Flexible Merging Control) Design the desired velocities $\{u_{i,x}^r, u_{i,y}^r\}, i\in\mathcal V,$
for each vehicle $i, i\in\mathcal V$ governed by \eqref{decouple_dynamics} such that all vehicles collectively merge into the platoon while maneuvering with arbitrary orderings, i.e., Objectives \ref{platoon_obj_1})-\ref{platoon_obj_4}) in Definition~\ref{def_platoon}.
\end{problem}

\begin{problem}
\label{problem_3}
(Velocity Execution) Calculate the actual inputs $\{v_i, \psi_i\}$ in \eqref{vehicle_dynamics} of vehicle $i$ for the execution based on the desired velocities $\{u_{i,x}^r, u_{i,y}^r\}, i\in\mathcal V,$ in Problems~\ref{problem_1} and \ref{problem_2}, i.e., $\{u_{i,x}^r, u_{i,y}^r\}  \rightarrow \{v_i, \psi_i\}$ according to \eqref{velocity_convertion}.
\end{problem}

A hierarchical architecture of Problems~\ref{problem_1}-\ref{problem_3} is given in Fig.~\ref{platoon_struture}, where Problems~\ref{problem_1}-\ref{problem_2} account for the upper-level signals, and Problem~\ref{problem_3} ensures the low-level tracking.


\begin{assumption}
\label{assum_set}
\cite{hu2024coordinated} For any $\kappa\in\mathbb{R}^+$ and the point $p_0$ outside $\mathcal T$ in~\eqref{target_set}, $\inf\{\phi(p_0): \mathrm{dist}(p_0, \mathcal T)\ge \kappa\}>0$ holds.
\end{assumption}

Assumption~\ref{assum_set} ensures the suitable selection of $\phi$ such that $\lim_{t \rightarrow \infty} \phi(p_0(t)) \leq 0 \Rightarrow \lim_{t \rightarrow \infty} \mathrm{dist}(p_0(t), \mathcal{T}) = 0$ for a point $p_0$ and $\mathcal{T}$ in~\ref{target_set}, thereby preventing pathological situations (see Example 3 in \cite{yao2018robotic}).

\begin{assumption}
\label{assump_yaw_angle}
The initial values of $\theta_i, \psi_i$ in \eqref{vehicle_dynamics} for each vehicle $i$ satisfy $\theta_i(0)=0, \psi_i(0)=0, \forall i\in\mathcal V$.
\end{assumption}

Assumption~\ref{assump_yaw_angle} stipulates that all the vehicles maneuver on the initial lanes at the beginning, which is necessary for lateral-lane keeping in Definition~\ref{def_pre_merging}.

\begin{assumption}
\label{assump_different_lane}
The initial vehicles on different lanes are assumed to be not overlapped in the longitudinal direction, i.e., $|x_{i}(0)-x_k(0)|>0, \{\forall i\neq k~|~y_i\neq y_k \}$.
\end{assumption}

Assumption~\ref{assump_different_lane} is necessary for vehicles on different lanes to determine whether to speed up or slow down, otherwise, they may get stuck in an overlapped dilemma.

\begin{assumption}
\label{assum_radius}
The initial positions of the vehicles on the same lane are larger than the vehicle base $B_i$, i.e., $|x_{i}(0)-x_{l}(0)|\geq r>B_i, \{\forall i\neq l~|~y_i=y_l\}$.
\end{assumption}


\section{C2TE Optimization Design}
In this section, since Problem \ref{problem_3} is achieved directly via~\eqref{velocity_convertion}, we will focus on the C2TE optimization algorithm design for Problems~\ref{problem_1}-\ref{problem_2}, respectively.

\subsection{Stage 1: Pre-Merging Regulation Control}

Suppose that all vehicles $\mathcal V$ move towards the positive direction of the x-axis. Let $i^{+}$ be the pre-sensing neighbor set of vehicle $i$ below,
\begin{equation}
\label{front_neighbor}
i^{+}=\left\{
\begin{aligned}
\mathrm{arg~min}_k\{x_k-x_{i}\}, & &R\geq x_k-x_{i}>0, k\neq i\in\mathcal V,\\
\emptyset, & &\mathrm{otherwise,}
\end{aligned}
\right.
\end{equation}
where $R$ is given in Definition~\ref{def_pre_merging}.
From \eqref{front_neighbor}, the cardinality of $i^{+}$ is an integer that contains at most one vehicle, i.e., $|i^+|\leq1$, which implies that $x_{i^+}$ is available to vehicle $i$ if $0<x_{i^+}-x_{i}\leq R$ (i.e., $i\neq \emptyset$).  Analogous to $\mathcal T$ in \eqref{target_set}, suppose that the longitudinal-regulation subtask $\mathcal T_{i,r}^{[1]}$ for vehicle $i$ and its pre neighbor $i^{+}\neq \emptyset$ is
\begin{align}
\label{stage_1_longitudinal_regulation0}
 \mathcal T_{i,r}^{[1]}=& \{\sigma_{i}\in\mathbb{R}~|~\phi_{i,r}^{[1]}(\sigma_i):=\nonumber\\
&(|\sigma_i-x_{i^+}|-R)^2\leq0\}, i^{+}\neq \emptyset,
\end{align}
with $\sigma_i$ being a dummy coordinate, and the symbol $[1]$ denoting Stage 1. For the case of $i^{+}=\emptyset$ in \eqref{front_neighbor}, $x_{i^+}$ is not utilized in \eqref{stage_1_longitudinal_regulation0}.
From the fact $x_{i^+}>x_i$ in \eqref{front_neighbor}, the target set $\mathcal T_{i,r}^{[1]}$ in \eqref{stage_1_longitudinal_regulation0} only contains a desired point at $\{x_{i^+}-R\}$. Substituting $x_i$ for $\sigma_i$ in \eqref{stage_2_lateral_converge} yields the longitudinal-regulation errors: $\phi_{i,r}^{[1]}(x_i):=(x_{i^+}-x_i-R)^2, i^{+}\neq \emptyset.$
Then, it follows from \eqref{decouple_dynamics} that the longitudinal-regulation CBF constraints are
\begin{align}
\label{stage_1_longitudinal_regulation_CBF}
\frac{\partial \phi_{i,r}^{[1]}(x_i)}{\partial x_i}u_{i,x}^r\leq -\gamma\big(\phi_{i,r}^{[1]}(x_i)\big)+\delta_{i,r}^{[1]},  i^{+}\neq \emptyset,
\end{align}
where $\gamma(\cdot)$ is the extended class $\mathcal K$ functions in Definition~\ref{definition_constraint_based}, and $\delta_{i,r}^{[1]}\in\mathbb{R}^+$ is the slack variable. 
It is observed in \eqref{stage_1_longitudinal_regulation0} and \eqref{stage_1_longitudinal_regulation_CBF} that there is no need to establish the longitudinal-regulation subtask $\mathcal{T}_{i,r}^{[1]}$ and the corresponding CBF constraint for vehicle $i$ when $i^+ = \emptyset$, as the condition $x_{i^+} - x_i > R$ in \eqref{front_neighbor} already provides sufficient longitudinal merging space for Stage 2.


Another subtask is to avoid collisions among vehicles on same lane. Let $\mathcal N_i^s$ be the same-lane sensing neighbor set,
\begin{align}
\label{same_lane_neighbor}
\mathcal N_i^s:=\{l\neq i\in\mathcal V~\big|~|x_{i}-x_l|\leq R~\mathrm{and}~y_i=y_l\}.
\end{align}
Here, only neighboring vehicles on the same lane are considered to possibly collide with vehicle $i$ in Stage~1. Since $x_i(t), x_l$ are time-varying, one has that $\mathcal N_i^s$ is time-varying, so is the communication topology in Stage 1. Then, the corresponding task set $\mathcal T_{i,l}^{[1]}$ between vehicles $i, l$ is characterized by
\begin{align}
\label{stage_1_same_lane_collision_avoidance}
\mathcal T_{i,l}^{[1]}=&\{\sigma_{i}\in\mathbb{R}~|~ \phi_{i,l}^{[1]}(\sigma_i):=r-|\sigma_i-x_l|\leq 0\}, l\in\mathcal N_i^s,
\end{align}
with $r$ being the safe distance in Definition~\ref{def_pre_merging}. Substituting $x_i$ for $\sigma_i$ in \eqref{stage_1_same_lane_collision_avoidance} yields the same-lane collision avoidance errors to be $\phi_{i,l}^{[1]}(x_i)=r-|x_i-x_l|, l\in\mathcal N_i^s$,
which follows from \eqref{decouple_dynamics} that the same-lane collision avoidance CBF constraints are formulated to be
\begin{align}
\label{stage_1_same_lane_collision_avoidance_CBF}
\frac{\partial \phi_{i,l}^{[1]}(x_i)}{\partial x_i}u_{i,x}^r\leq -\gamma\big(\phi_{i,l}^{[1]}(x_i)\big).
\end{align}
By combining Definition~\ref{definition_constraint_based}, the CBF constraints \eqref{stage_1_longitudinal_regulation_CBF} and \eqref{stage_1_same_lane_collision_avoidance_CBF} together, we formulate the C2TE algorithm (pre-merging regulation in Stage 1) for vehicle $i, i\in\mathcal V,$  to be a distributed constraint-based optimization one,
\begin{subequations}
\label{MR_optimization_1}
\begin{align}
&\min\limits_{u_{i,x}^r,\delta_{i,r}^{[1]}} \Big\{(u_{i,x}^r)^2+c(\delta_{i,r}^{[1]})^2\Big\}\\
\mathrm{s.t.}~&\textbf{C1}:  \mathrm{Longitudinal~regulation~in~} \eqref{stage_1_longitudinal_regulation_CBF},\\
&\textbf{C2}: \mathrm{Same}~\mathrm{lane~collision~avoidance~in~}\eqref{stage_1_same_lane_collision_avoidance_CBF}
\end{align}		
\end{subequations}
with $c\in\mathbb{R}^+$ being the weight of slack variables. Although the introduction of $c\in \mathbb{R}^+$ in (\ref{MR_optimization_1}a) adds complexity compared to Definition~\ref{definition_constraint_based}, its primary function is to regulate the convergence speed. Moreover, it has been shown in (5) of \cite{notomista2019optimal} that the constraint-based task execution, similar to Definition~\ref{definition_constraint_based}, remains effective with the inclusion of $c$. The cost function (\ref{MR_optimization_1}a) contains two terms, where the first term $(u_{i,x}^r)^2$ accounts for the minimum-energy execution, and the second term $(\delta_{i,r}^{[1]})^2$ is to minimize the constraint violation  (\ref{MR_optimization_1}b). From the interaction of \textbf{C1}-\textbf{C2}, all the vehicles will be proved to safely converge to sufficient longitudinal distances later. 
Moreover, from the optimization \eqref{MR_optimization_1} in Stage 1, one has that $u_{i,x}^r$ may be negative (i.e., $u_{i,x}^r\leq 0$), which inevitably results in reversing or sharp turning. To avoid such issues, we add a pre-designed cruising velocity $v_d$ to each vehicle, i.e., $\{u_{i,x}^r+v_d\}$ with $v_d\in\mathbb{R}^+$ being the desired velocity given in \eqref{virtual_target_dynamics} later. Therefore, all vehicles can dynamically regulate their longitudinal distances.


\begin{remark}
The proposed C2TE algorithm~\eqref{MR_optimization_1} can still increase longitudinal spacing among vehicles by allowing the rear vehicle to move backward, away from other vehicles, even when the initial longitudinal space is limited. By leveraging the CBF constraints \eqref{stage_1_longitudinal_regulation_CBF}, each vehicle $i, i \in \mathcal{V}$ can adjust its position to increase the longitudinal distance from its pre-sensing neighbor $i^+$ in \eqref{front_neighbor}. However, if there is not enough space for vehicle $i$ to move backward, the rear vehicle which has no restriction on backward movement, can create enough longitudinal space. This allows the vehicle ahead of the rear one to move backward as well. Iteratively, all the vehicles finally gain sufficient longitudinal spacing, 
which necessarily requires on-board communication equipment.
\end{remark}

\subsection{Stage 2: Ordering-Flexible MVP Merging Control}


Suppose the lateral-convergence subtask $\mathcal T_{i,c}^{[2]}$ for vehicle~$i$ is characterized to be
\begin{align}
\label{stage_2_lateral_converge}
\mathcal T_{i,c}^{[2]}=&\{\sigma_{i}\in\mathbb{R}~|~\phi_{i,c}^{[2]}(\sigma_i):=|\sigma_i-y_d|\leq0\}, i\in\mathcal V,
\end{align}
with $\sigma$ being the dummy coordinate, the label $[2]$ denoting Stage 2 and $y_d$ given in \eqref{vehicle_i_desired_lane}. Substituting $y_i$ in \eqref{vehicle_dynamics} into \eqref{stage_2_lateral_converge} yields the lateral-convergence errors $\phi_{i,c}^{[2]}(y_i)=|y_i-y_d|$,
which implies that the lateral-convergence CBF constraints for $\mathcal T_{i,c}^{[2]}$ are 
\begin{align}
\label{stage_2_lateral_converge_CBF}
\frac{\partial \phi_{i,c}^{[2]}(y_i)}{\partial y_i}u_{i,y}^r\leq -\gamma\big(\phi_{i,c}^{[2]}(y_i)\big)+\delta_{i,c}^{[2]},
\end{align}
with $u_{i,y}^r$ in \eqref{decouple_dynamics} and $\delta_{i,c}^{[2]}\in\mathbb{R}^+$ being a slack variable.

To achieve the longitudinal-target attraction subtasks, we define a virtual target vehicle below,
\begin{align}
\label{virtual_target_dynamics}
\dot{x}_d=&v_d,~\dot{y}_d=0,
\end{align}
where $p_d:=[x_d, y_d]\t\in\mathbb{R}^2, v_d\in\mathbb{R}^+$ are the position, and velocity, respectively. 
Then, we suppose the longitudinal-target attraction subtask $\mathcal T_{i,m}^{[2]}$ between vehicle~$i, i\in\mathcal V,$ and the virtual target $p_d$ in~\eqref{virtual_target_dynamics} is 
\begin{align}
\label{stage_2_longitudinal_merging}
\mathcal T_{i,m}^{[2]}=\{\sigma_{i}\in\mathbb{R}~|~\phi_{i,m}^{[2]}(\sigma_i):=|\sigma_i-x_d|\leq 0\}.
\end{align}
Substituting $x_i$ for $\sigma_i$ in \eqref{stage_2_longitudinal_merging} yields the longitudinal target-attraction errors to be $\phi_{i,m}^{[2]}(x_i)=|x_i-x_d|$,
which further implies that the longitudinal target-attraction CBF constraints are
\begin{align}
\label{stage_2_longitudinal_merging_CBF}
\frac{\partial \phi_{i,m}^{[2]}(x_i)}{\partial x_i}(u_{i,x}^r-v_d)\leq -\gamma\big(\phi_{i,m}^{[2]}(x_i)\big)+\delta_{i,m}^{[2]}
\end{align}
with $\delta_{i,m}^{[2]}\in\mathbb{R}^+$ being a slack variable, $u_{i,x}^r$ and $v_d$ are given in \eqref{decouple_dynamics} and \eqref{virtual_target_dynamics}, respectively. Here, the constraint~\eqref{stage_2_longitudinal_merging_CBF} includes the additional $v_d$, which not only aims to track the target's position $x_d$ but also tracks the target's velocity $v_d$.
Let $\mathcal N_i$ be the sensing neighbor set of vehicle $i$ in the longitudinal direction below,
\begin{align}
\label{neighbor_set}
\mathcal N_i:=\{j\neq i\in\mathcal V~\big|~|x_{i}-x_j|\leq R\}.
\end{align}
Note that $\mathcal N_i^s$ is time-varying, so is the communication topology in Stage 2. We define the neighboring collision avoidance subtasks $\mathcal T_{i,j}^{[2]}$ between vehicles $i$ and $j$ in the longitudinal direction to be
\begin{align}
\label{stage_2_longitudinal_collision_avoidance}
\mathcal T_{i,j}^{[2]}=&\{\sigma_{i}\in\mathbb{R}~|~\phi_{i,j}^{[2]}(\sigma_i):=\frac{1}{|\sigma_i-x_j|-r}-\frac{1}{\rho-r}\leq 0\},\nonumber\\
				&i\in\mathcal V, j\in\mathcal N_i,
\end{align}
with a constant $\rho\in(r, R)$ in Definition \ref{def_pre_merging}.
Substituting $x_i$ for $\sigma_i$ in \eqref{stage_2_longitudinal_collision_avoidance} yields the neighboring collision avoidance errors
\begin{align}
\label{stage_2_longitudinal_collision_avoidance_err}
\phi_{i,j}^{[2]}(x_i)=\frac{1}{|x_i-x_j|-r}-\frac{1}{\rho-r},
\end{align}
where $\phi_{i,j}^{[2]}(x_i)\leq 0$ if $|x_i-x_j|\geq\rho$, and $\phi_{i,j}^{[2]}(x_i)\rightarrow+\infty$ if $|x_i-x_j|\rightarrow r$. The collision-avoidance errors $\phi_{i,j}^{[2]}(x_i)$ are different from  \eqref{stage_1_same_lane_collision_avoidance_CBF}, which not only guarantee the collision avoidance but also will be utilized to achieve the {\it ordering-flexible platoon} in Lemmas~\ref{stage_2_lemma_longitudinal_coolision}-\ref{stage_2_lemma_longitudinal_ordering} later.
The neighboring collision avoidance CBF constraints become
\begin{align}
\label{stage_2_longitudinal_collision_avoidance_CBF}
\frac{\partial \phi_{i,j}^{[2]}(x_i)}{\partial x_i}(u_{i,x}^r-v_d)\leq -\gamma\big(\phi_{i,j}^{[2]}(x_i)\big)+\delta_{i,j}^{[2]}
\end{align}
with the slack variables $\delta_{i,j}^{[2]}\in\mathbb{R}^+$. Analogously, the counterparts $u_{i,x}^r - v_d$ are incorporated into \eqref{stage_2_longitudinal_collision_avoidance_CBF}, which is to align with \eqref{stage_2_longitudinal_merging_CBF} and eliminate the influence of $v_d$ on collision avoidance missions.
According to Definition~\ref{definition_constraint_based}, it, together with \eqref{stage_2_lateral_converge_CBF}, \eqref{stage_2_longitudinal_merging_CBF}, \eqref{stage_2_longitudinal_collision_avoidance_CBF}, gives that the C2TE algorithm ({\it ordering-flexible} platoon merging in Stage 2) is a distributed constraint-based optimization one,
\begin{subequations}
\label{MR_optimization_2}
\begin{align}
&\min\limits_{u_{i,x}^r,u_{i,y}^r, \delta_{i,m}^{[2]}, \delta_{i,c}^{[2]},\delta_{i,j}^{[2]}} \Big\{(u_{i,x}^r-v_d)^2+(u_{i,y}^r)^2\nonumber\\
&~~~~~~~~~~~+c\Big[(\delta_{i,m}^{[2]})^2+(\delta_{i,c}^{[2]})^2+\sum_{j\in\mathcal N_i}(\delta_{i,j}^{[2]})^2\Big]\Big\}\\
\mathrm{s.t.}~&\textbf{C3}:  \mathrm{Lateral~convergence~in~}\eqref{stage_2_lateral_converge_CBF},\\
&\textbf{C4}: \mathrm{Virtual~target~attraction~in~}\eqref{stage_2_longitudinal_merging_CBF},\\
&\textbf{C5}: \mathrm{Neighbor~collision~avoidance~in}~\eqref{stage_2_longitudinal_collision_avoidance_CBF},
\end{align}		
\end{subequations}
with $c\in\mathbb{R}^+$ being the weight to regulate the convergence speed as well (i.e., the control gain of the optimal inputs \eqref{stage2_lateral_input1} and \eqref{stage2_long_input2_case3} later). The cost function (\ref{MR_optimization_2}a) contains five terms, where the first two terms $\{(u_{i,y}^r)^2+(u_{i,x}^r-v_d)^2\}$ account for Objective \ref{platoon_obj_3}) in Definition~\ref{def_platoon}, and the last three terms $\{c[(\delta_{i,m}^{[2]})^2+(\delta_{i,c}^{[2]})^2+\sum_{j\in\mathcal N_i}(\delta_{i,j}^{[2]})^2]\}$ are to minimize the violation of \textbf{C3}-\textbf{C5}. \textbf{C3} and \textbf{C5} account for Objectives~\ref{platoon_obj_1}), \ref{platoon_obj_4}) of Definition~\ref{def_platoon}, respectively. By interacting \textbf{C4}-\textbf{C5} together, the platoon is formed with arbitrary orderings, i.e., Objective \ref{platoon_obj_2}) in Definition~\ref{def_platoon}.

\begin{remark}
By incorporating the cost term $(u_{i,x}^r - v_d)^2$ in \eqref{MR_optimization_2} along with $(u_{i,x}^r - v_d)$ in the longitudinal target-attraction CBF constraint \eqref{stage_2_longitudinal_merging_CBF}, the velocity of each vehicle $i$ can eventually converge to $v_d$ (i.e., $v_i \rightarrow v_d$ for $i \in \mathcal{V}$), thereby achieving platoon cruising. Conversely, if $v_d$ is removed from \eqref{stage_2_longitudinal_merging_CBF}, a platoon may not form because the virtual target $x_d$ will continue to move. Additionally, by aligning $(u_{i,x}^r - v_d)$ in both constraints \eqref{stage_2_longitudinal_merging_CBF} and \eqref{stage_2_longitudinal_collision_avoidance_CBF}, the interaction between the attraction from the longitudinal target-attraction constraint  \eqref{stage_2_longitudinal_merging_CBF}  and the repulsion from the neighboring collision avoidance constraints \eqref{stage_2_longitudinal_collision_avoidance_CBF} will reach a balance, thereby achieving an order-flexible platoon, i.e., Objective $(2)$ of Definition \ref{def_platoon}.
\end{remark}

\begin{remark}
\label{remark_slack_argument}
Different from the collision-avoidance constraints~$\textbf{C2}$~in (\ref{MR_optimization_1}c) which contain no slack variables to achieve collision avoidance via forward invariance \cite{notomista2019constraint,ames2016control} later,  the constraints $\textbf{C5}$ in~(\ref{MR_optimization_2}d) introduce an additional slack variable $\delta_{i,j}^{[2]}$ to align with the constraint $\textbf{C4}$, which guarantees the strong duality, and thus can be utilized to prove the {\it ordering-flexible} MVP merging in Lemma~\ref{stage_2_lemma_longitudinal_ordering}. However, the introduction of such slack variables may lead to the undesirable issue where the inter-vehicle distances violate the desired collision avoidance (i.e., $|x_i-x_j|>\rho\Leftrightarrow x_i\notin \mathcal T_{i,j}^{[2]}$). 
Therefore, one may not be able to prove collision avoidance using traditional forward invariance \cite{notomista2019constraint,ames2016control}. Nevertheless, we will show later that collision avoidance is still guaranteed with a new proof. In other words, we leverage the boundedness of a Lyapunov function to show that no collision will happen. Technical details refer to Appendix \ref{Appendix_B} later.
\end{remark}

\begin{remark}
The information for each vehicle at each stage is given below. (i) For the pre-merging regulation in Stage~1, it follows from \eqref{MR_optimization_1} that each vehicle $i \in \mathcal{V}$ has access to its longitudinal position $x_i$, the pre-sensing neighbor $x_{i^+}$, the same-lane sensing neighbor set $x_l, l \in \mathcal{N}_i^s$, and the sensing and collision radii $R$ and $r$. These elements are used to establish the longitudinal regulation and same-lane collision avoidance CBF constraints in \eqref{stage_1_longitudinal_regulation_CBF} and \eqref{stage_1_same_lane_collision_avoidance_CBF}. (ii) For the {\it ordering-flexible platoon} merging in Stage~2, it follows from \eqref{MR_optimization_2} that each vehicle $i \in \mathcal{V}$ has access to its longitudinal and lateral positions $[x_i, y_i]^T$, the positions and velocities of the virtual target $[x_d, y_d]^T$ and $[v_d, 0]^T$, the longitudinal positions of the sensing neighbor set $x_j, j \in \mathcal{N}_i$, as well as $R$ and $r$. These elements will be utilized to establish the lateral convergence, virtual-target attraction, and neighbor collision avoidance CBF constraints in \eqref{stage_2_lateral_converge_CBF}, \eqref{stage_2_longitudinal_merging_CBF}, and \eqref{stage_2_longitudinal_collision_avoidance_CBF}.
\end{remark}

\begin{remark}
Compared to the traditional CBF algorithms \cite{notomista2019constraint,ames2016control}, two theoretical challenges arise in the {\it ordering-flexible} platoon merging mission. (i) The first challenge is the additional slack variable $\delta_{i,j}^{[2]}$ in collision-avoidance CBF constraints $\textbf{C5}$ in~(\ref{MR_optimization_2}d). This may lead to an undesirable situation where inter-vehicle distances cannot be proved using the traditional forward invariance \cite{notomista2019constraint,ames2016control}. Technical details refer to Remark~\ref{remark_slack_argument} and Appendix~\ref{Appendix_B}. (ii) The second challenge is the strong nonlinear couplings introduced by the interaction between longitudinal-target attraction and the time-varying neighboring collision avoidance constraints $\textbf{C4-C5}$ in~(\ref{MR_optimization_2}c)-(\ref{MR_optimization_2}d). This makes the rigorous convergence analysis of the {\it ordering-flexible} platoon sophisticated to analyze. To tackle this issue, we divide the convergence of platoon cruising and flexible ordering into four cases and three claims, respectively, and use the contradiction. Technical details refer to Appendix~\ref{Appendix_C} and~\ref{Appendix_D} later.
\end{remark}

\section{Feasibility and Convergence Analysis}

\subsection{Feasibility Guarantee}
\label{sec_feasible}
Recalling Assumptions~\ref{assump_different_lane}-\ref{assum_radius}, we ensure that there always exists a trivial feasible solution for vehicle $i$ in (\ref{MR_optimization_1}), i.e., $\big\{u_{i,x}^{r}=v_d, \delta_{i,l}^{[1]} \mbox{ is sufficiently large}\big\},$
which satisfies \textbf{C1}-\textbf{C2} in (\ref{MR_optimization_1}b)-(\ref{MR_optimization_1}c) due to the arbitrary selection of the slack variable $\delta_{i,l}^{[1]} $.
Analogously, we ensure that there always exists a trivial feasible solution for vehicle $i$ in  (\ref{MR_optimization_2}), i.e., $\big\{u_{i,x}^r=v_d, u_{i,y}^r=0, \delta_{i,c}^{[2]}, \delta_{i,m}^{[2]}, \delta_{i,j}^{[2]}, j\in\mathcal N_i, \mbox{ are sufficiently large}\big\},$
which satisfies \textbf{C3}-\textbf{C5} in (\ref{MR_optimization_2}b)-(\ref{MR_optimization_2}d) due to the selection of $\delta_{i,c}^{[2]}, \delta_{i,m}^{[2]}, \delta_{i,j}^{[2]}$.

Note that the sufficiently large $\delta_{i,l}^{[1]}, \delta_{i,c}^{[2]}$, $\delta_{i,m}^{[2]}$, $\delta_{i,j}^{[2]}$ in (\ref{MR_optimization_1}) and (\ref{MR_optimization_2}) will not disrupt the convergence of the tracking errors. Firstly, the sufficiently large slack variables are only to ensure the solutions for (\ref{MR_optimization_1}) and (\ref{MR_optimization_2}) at the initial stage. However, once using KKT conditions, the optimal slack variables immediately become associated with the CBF tracking errors. For instance, $\delta_{i,c}^{[2]\ast}$
in Case~2 of Appendix~\ref{Appendix_A} is only determined by the CBF erros $\phi_{i,c}^{[2]}(y_i)$. 
Moreover, the optimal slack variables do not affect the optimal inputs for (\ref{MR_optimization_1}) and (\ref{MR_optimization_2}) as well. According to the KKT conditions, the $i$-th optimal inputs $u_{i,x}^{r\ast}$ and $u_{i,y}^{r\ast}$ depend on the CBF tracking errors rather than the slack variables (see examples in \eqref{stage2_lateral_input1} and \eqref{stage2_long_input2_case3} later).
Consequently, the CBF tracking errors are only regulated by the optimal inputs $u_{i,x}^{r\ast}$ and $u_{i,y}^{r\ast}$.

\subsection{Pre-Merging Regulation Convergence}
\label{sec_pre_merging_convergence}


Since the lateral-lane keeping in Objective \ref{Pre_platoon_obj_2}) of Definition~\ref{def_pre_merging} is naturally satisfied under Assumption~\ref{assump_yaw_angle}, we will prove Objectives \ref{Pre_platoon_obj_3}), \ref{Pre_platoon_obj_1}) of Definition~\ref{def_pre_merging}, respectively.

\begin{lemma}
\label{stage_1_same_lane}
All vehicles $\mathcal V$ governed by \eqref{decouple_dynamics} and \eqref{MR_optimization_1} achieve Objective~\ref{Pre_platoon_obj_3}) of Definition~\ref{def_pre_merging}. Namely, $|x_{i}(t)-x_l(t)|\geq r, \forall t\geq0, \{ i\neq l~|~y_i=y_l\}$.
\end{lemma}

{\it Proof.} 
It follows from \cite{ames2016control} that the optimization \eqref{MR_optimization_1} is a typical control-Lyapunov-function-control-barrier-function quadratic program (CLF-CBF-QP) problem, which inherits the 
forward invariance in~\eqref{property}.
Recalling $\mathcal T_{i,l}^{[1]}$ and $\phi_{i,l}^{[1]}(x_i)$ in \eqref{stage_1_same_lane_collision_avoidance} and \eqref{stage_1_same_lane_collision_avoidance_CBF}, one has that
${\partial \phi_{i,l}^{[1]}(x_i)}/{\partial x_i}=-{x_i-x_l}/{|x_i-x_l|}\neq0, \forall x_i\in \mathcal T_{i,l}^{[1]}$
because of $x_i\in \mathcal T_{i,l}^{[1]}\Leftrightarrow |x_i-x_l|>r>0$, which implies that 
${\partial \phi_{i,l}^{[1]}(x_i)}/{\partial x_i}$ is locally Lipschitz continuous. According to (Theorem 3 in \cite{ames2016control}), the optimal input $u_{i,x}^{r\ast}$ in \eqref{MR_optimization_1} is locally Lipschitz continuous as well. From Assumption~\ref{assum_radius} and \eqref{stage_1_same_lane_collision_avoidance},
the initial states of $\phi_{i,l}^{[1]}(x_i(0))=r-|x_i(0)-x_l(0)|\leq0$, which implies that $x_i(0)\in\mathcal{T}_{i,l}^{[1]}$. Using the forward invariance in \eqref{property},  it gives $x_i(t)\in\mathcal{T}_{i,l}^{[1]}, \forall t>0$. The proof is completed.
\eproof

\begin{lemma}
\label{stage_1_longitudinal_regulation}
Under Assumptions~\ref{assum_set} and \ref{assump_different_lane}, all vehicles $\mathcal V$ governed by \eqref{decouple_dynamics} and \eqref{MR_optimization_1} achieve Objective~\ref{Pre_platoon_obj_1}) of Definition~\ref{def_pre_merging}. Namely, $\lim_{t\rightarrow\infty}|x_{i}(t)-x_j(t)|\geq R, \forall i\neq j\in\mathcal V$.
\end{lemma}

{\it Proof.}
According to the pre-sensing neighbor $i^+$ in \eqref{front_neighbor}, we divide the convergence into three cases.


\textbf{Case 1:} $i^{+}=\emptyset, \forall t>0$. If Case 1 holds, it follows from the definition of $i^{+}$ in \eqref{front_neighbor} that $x_{i^+}(t) - x_i(t) > R$ holds for all $t > 0$, which implies that the asymptotic condition of $\lim_{t \rightarrow \infty} |x_{i^+}(t) - x_i(t)| \geq R$ in Case 1 is naturally satisfied. Furthermore, recalling that there is no longitudinal-regulation CBF constraint in \eqref{stage_1_longitudinal_regulation_CBF} for vehicle~$i$ when $i^{+} = \emptyset$, the asymptotic-convergence property in previous CBF works \cite{ames2016control,xu2015robustness} is not utilized in this case.

\textbf{Case 2:} $i^{+}\neq \emptyset, \forall t>0$. If Case 2 holds, one has that $x_{i^+}(t)-x_{i}(t)\leq R,  \forall t>0$, which implies that the constraint \eqref{stage_1_longitudinal_regulation_CBF} always exists.Analogous to Lemma~\ref{stage_1_same_lane}, it follows from \eqref{stage_1_longitudinal_regulation_CBF} and Assumption~\ref{assump_different_lane} that ${\partial \phi_{i,r}^{[1]}(x_i)}/{\partial x_i}=2(x_{i^+}-x_i-R)\neq 0, $
which is locally Lipschitz continuous. According to Theorem~11 in \cite{xu2015robustness}, the optimal inputs $u_{i,x}^{r\ast}(x_i)$ in \eqref{MR_optimization_1} are locally Lipschitz continuous for $x_i\notin\mathcal T_{i,r}^{[1]}$. Using the asymptotic convergence in  \eqref{property}, vehicle $i$ finally converges to $\mathcal T_{i,r}^{[1]}$, i.e., $\lim_{t\rightarrow\infty}\phi_{i,r}^{[1]}(x_i(t))=0$ because $\mathcal T_{i,r}^{[1]}$ only contains a point $x_{i^+}-R$, which implies that $\lim_{t\rightarrow\infty}x_{i^{+}}(t)-x_i(t)=R$ is proved. 

\textbf{Case 3:} $i^{+}=\emptyset$ or $i^{+}\neq \emptyset$. If Case 3 holds, the asymptotic condition of $\lim_{t \rightarrow \infty} |x_{i^+}(t) - x_i(t)| \geq R$ is naturally satisfied for $i^{+} = \emptyset$, as shown in Case 1. For $i^{+} \neq \emptyset$, this condition can also be derived from the asymptotic convergence in \eqref{property}, as shown in Case 2. From Cases 1-3, one has that $\lim_{t \rightarrow \infty} |x_{i^+}(t) - x_i(t)| \geq R$, which follows from $i^+$ that $\lim_{t\rightarrow\infty}|x_{i}(t)-x_j(t)|\geq R, \forall i\neq j\in\mathcal V$. The proof is thus completed.
\eproof

\begin{theorem}
\label{theorem_1}
Under Assumptions~\ref{assum_set}-\ref{assum_radius}, all vehicles $\mathcal V$ governed by \eqref{decouple_dynamics} and \eqref{MR_optimization_1} achieve the pre-merging regulation, i.e., Objectives~\ref{Pre_platoon_obj_1})- \ref{Pre_platoon_obj_3}) in Definition~\ref{def_pre_merging}. 
\end{theorem}

{\it Proof.}
We conclude Lemmas~\ref{stage_1_same_lane}-\ref{stage_1_longitudinal_regulation}.
\eproof

\subsection{Ordering-Flexible MVP Merging Convergence}

To facilitate the practical usage of $\lim_{t \rightarrow \infty} |x_i(t) - x_j(t)| \geq R, \forall i \neq j \in \mathcal{V}$ in Lemma~\ref{stage_1_longitudinal_regulation}, we specifically choose a constant $\rho \in (r, R)$ in \eqref{stage_2_longitudinal_collision_avoidance}, rather than $R$ as the stage switching value in the following Assumption. 

\begin{assumption}
\label{assumption_state_1_change}
Each vehicle $i, i\in\mathcal V$ governed by \eqref{MR_optimization_2} get sufficient longitudinal distance after a finite time $T_1>0$, i.e., $\lim_{t\rightarrow T_1}|x_{i}(t)-x_j(t)|\geq \rho, \forall j\neq i\in\mathcal V$.
\end{assumption}

Assumption~\ref{assumption_state_1_change} only requires the existence of a finite time $T_1$, which is convenient to verify. 
Specifically, $\mathrm{For}~\forall \rho<R, ~\exists T_1>0~\mathrm{such~that}~\lim_{t\rightarrow T_1}|x_{i}(t)-x_j(t)|\geq \rho, \forall j\neq i\in\mathcal V.$
Under Assumption~\ref{assumption_state_1_change}, each vehicle $i$ essentially depends on whether the minimum longitudinal distance between itself and the other vehicle exceeds $\rho$. 
Since $u_{i,x}^r, \delta_{i,c}^{[2]}$ and $u_{i,y}^r, \delta_{i,m}^{[2]}, \delta_{i,j}^{[2]}$ are determined by (\ref{MR_optimization_2}c)-(\ref{MR_optimization_2}d) and  (\ref{MR_optimization_2}b), respectively, the optimization (\ref{MR_optimization_2}) isdecoupled into the longitudinal and lateral optimization ones.

\textbf{Lateral optimization:}
\begin{subequations}
\label{lateral_MR_optimization_2}
\begin{align}
&\min\limits_{u_{i,y}^r, \delta_{i,c}^{[2]}} \Big\{(u_{i,y}^r)^2+c(\delta_{i,c}^{[2]})^2\Big\}\\
\mathrm{s.t.}~&\textbf{C3}:  \mathrm{Lateral~convergence~in~}\eqref{stage_2_lateral_converge_CBF}.
\end{align}		
\end{subequations}
\textbf{Longitudinal optimization:}
\begin{subequations}
\label{longitudinal_MR_optimization_2}
\begin{align}
&\min\limits_{u_{i,x}^r,\delta_{i,m}^{[2]}, \delta_{i,j}^{[2]}} \Big\{(u_{i,x}^r-v_d)^2+c\Big[(\delta_{i,m}^{[2]})^2+\sum_{j\in\mathcal N_i}(\delta_{i,j}^{[2]})^2\Big]\Big\}\\
\mathrm{s.t.}~& \textbf{C4}: \mathrm{Virtual~target~attraction~in~}\eqref{stage_2_longitudinal_merging_CBF},\\
&\textbf{C5}: \mathrm{Neighbor~collision~avoidance~in}~\eqref{stage_2_longitudinal_collision_avoidance_CBF}.
\end{align}		
\end{subequations}

\begin{lemma}
\label{stage_2_lemma_lateral}
Under Assumption~\ref{assum_set}, all vehicles $\mathcal V$ governed by \eqref{decouple_dynamics} and \eqref{lateral_MR_optimization_2} achieve Objective~\ref{platoon_obj_1}) of Definition~\ref{def_platoon} with a simple extended class $\mathcal K$ function $\gamma(\phi)=\phi$, i.e., $\lim_{t\rightarrow\infty}y_i(t)=y_d, \forall i\in \mathcal V$.
\end{lemma}

{\it Proof.}
See Appendix~\ref{Appendix_A}.
\eproof


\begin{lemma}
\label{stage_2_lemma_longitudinal_coolision}
All vehicles $\mathcal V$ governed by \eqref{decouple_dynamics} and \eqref{longitudinal_MR_optimization_2} achieve Objective~\ref{platoon_obj_4}) of Definition~\ref{def_platoon} with a simple extended class $\mathcal K$ function $\gamma(\phi)=\phi$, i.e., $|x_i(t)-x_j(t)|\geq r, \forall i\neq j\in\mathcal V,  t\geq0$.
\end{lemma}

{\it Proof.}
See Appendix~\ref{Appendix_B}.
\eproof

\begin{lemma}
\label{stage_2_lemma_longitudinal_crusing}
All vehicles $\mathcal V$ governed by \eqref{decouple_dynamics} and \eqref{longitudinal_MR_optimization_2} achieve Objective~\ref{platoon_obj_3}) of Definition~\ref{def_platoon}, i.e, $\lim_{t\rightarrow\infty}\dot{x}_i(t)$ $=\lim_{t\rightarrow\infty}\dot{x}_j(t)=v_d, \forall i\neq j\in\mathcal V$.
\end{lemma}

{\it Proof.}
See Appendix~\ref{Appendix_C}.
\eproof

\begin{lemma}
\label{stage_2_lemma_longitudinal_ordering}
All vehicles $\mathcal V$ governed by \eqref{decouple_dynamics} and \eqref{longitudinal_MR_optimization_2} achieve Objective~\ref{platoon_obj_2}) of Definition~\ref{def_platoon}, i.e., $r<\lim_{t\rightarrow\infty}$ $|x_{s[k]}(t)-x_{s[k+1]}(t)|<\rho, k\in\mathbb{Z}_1^{N-1}$.
\end{lemma}

{\it Proof.}
See Appendix~\ref{Appendix_D}.
\eproof

\begin{theorem}
\label{theorem_platoon_merging}
Under Assumption~\ref{assum_set} and \ref{assumption_state_1_change}, all vehicles $\mathcal V$ governed by \eqref{decouple_dynamics} and \eqref{MR_optimization_2} achieve the {\it ordering-flexible} platoon merging, i.e., Objectives~\ref{platoon_obj_1})- \ref{platoon_obj_4}) in Definition~\ref{def_platoon}. 
\end{theorem}

{\it Proof.}
We conclude from Lemmas \ref{stage_2_lemma_lateral}-\ref{stage_2_lemma_longitudinal_ordering}.
\eproof

\begin{remark}
The enlarged longitudinal distance governed by the C2TE algorithm \eqref{MR_optimization_1} in Stage~1 may reduce the fusion efficiency and road throughput if $\rho$ in Definition~\ref{def_pre_merging} is set to be too large. $\mathrm{(i)}$ On the one hand, by shrinking $\rho$ while satisfying $\rho>r$, one can speed up Stage 1 to increase efficiency.  On the other hand, the enlarged longitudinal distance only exists for a short time for each vehicle $i$. Once each vehicle $i$ achieves sufficient longitudinal separation from other vehicles, it transitions to Stage 2, rapidly reducing the enlarged longitudinal distance while ensuring collision avoidance (i.e., $|x_i(t) - x_{j}(t)| \geq r, \forall i\neq j\in\mathcal V, t>0$) with $\rho>r$. $\mathrm{(ii)}$ The two-phase merging (i.e., Stages 1-2) proceeds sequentially for each vehicle, which also maintains high merging efficiency, even in large-scale scenarios.  Additional evidence will be provided in Fig.~\ref{different_lane_1} later. $\mathrm{(iii)}$ The efficiency of the C2TE algorithms \eqref{MR_optimization_1} and \eqref{MR_optimization_2} essentially lies in a flexible platoon with arbitrary ordering, which reduces the overall movement trajectories compared to a fixed-order platoon. For instance, if the initial position of vehicle $i$ is at the end of the group but its desired position is predefined at the front, vehicle $i$ will inevitably travel longer trajectories, resulting in lower efficiency. In contrast, such unnecessary movements can be avoided using {\it ordering-flexible} platoon in this paper. 
\end{remark}

\begin{remark}
For efficient finite-time convergence which has been explored in prior CBF works (Proposition 7 in \cite{notomista2019constraint}), the C2TE algorithms \eqref{MR_optimization_1} and \eqref{MR_optimization_2} also achieve it to some extent by replacing~$\gamma(x)=c(x)^\eta$ with $c > 0$ and $\eta \in (0,1)$. $\mathrm{(i)}$ In Stage 1, by replacing $\gamma(\phi_{i,r}^{[1]}(x_i))$ with $c(\phi_{i,r}^{[1]}(x_i))^\eta$ in the longitudinal-regulation constraints \eqref{stage_1_longitudinal_regulation_CBF}, 
the C2TE algorithm \eqref{MR_optimization_1} increases the efficiency. However, due to the same-lane collision avoidance constraints in \eqref{stage_1_same_lane_collision_avoidance_CBF}, the finite-time convergence in Stage 1 may inevitably be delayed. $\mathrm{(ii)}$~In Stage~2, for the lateral optimization~\eqref{lateral_MR_optimization_2}, the finite-time convergence is achieved by replacing~$\gamma(\phi_{i,c}^{[2]}(y_i)):=c(\phi_{i,c}^{[2]}(y_i))^\eta$ in \eqref{stage_2_lateral_converge_CBF},
which follows from \eqref{stage_2_longitudinal_merging_CBF} that the finite time of the lateral convergence is calculated by 
$T=\max\{\frac{2}{c(1-\eta)}\phi_{i,c}^{[2]}(y_i(0))^{1-\eta}, \forall i\in\mathcal V\}$ ~\cite{notomista2019constraint}.  
For the longitudinal optimization \eqref{longitudinal_MR_optimization_2}, due to the balance condition of \eqref{Lemma5_case_4_convergence} between the virtual-target attraction and the neighbor collision avoidance constraints, the function $\gamma(\cdot)$ cannot be changed in \eqref{stage_2_longitudinal_merging_CBF} and \eqref{stage_2_longitudinal_collision_avoidance_CBF}. However, since the Objective 2) in Definition~\ref{def_platoon} is roughly stipulated by $r<\lim_{t\rightarrow\infty}|x_{s[k]}(t)-x_{s[k+1]}(t)|<\rho, k\in\mathbb{Z}_1^{N-1}$, we can also reduce the convergence time by increasing $\rho$. Finally,  formal analysis of the finite-time MVP merging mission is still challenging and requires further investigation.
\end{remark}

\begin{remark}
The C2TE algorithms \eqref{MR_optimization_1} and~\eqref{MR_optimization_2} are adaptable to an arbitrary number of lanes and arbitrary choice of the desired lane.  Firstly, recalling Stage 1, all vehicles governed by \eqref{MR_optimization_1} only regulate their longitudinal distances on their initial lanes, which ensures that collision avoidance within the same lane remains unaffected. Secondly, $\mathrm{(i)}$ for the lateral optimization~\eqref{lateral_MR_optimization_2} in Stage 2, the number of lanes and choice of desired lane affect only the initial lateral distance between each vehicle and the desired lane. $\mathrm{(ii)}$ For the longitudinal optimization~\eqref{longitudinal_MR_optimization_2} in Stage 2, the number of lanes and choice of desired lane do not influence the longitudinal direction as well. Last but not least, additional simulations with different numbers of lanes will be provided in Fig.~\ref{different_lane_1} later. 
\end{remark}

\begin{remark}
$\mathrm{(i)}$ For the extreme disturbance of vehicles breaking down, the C2TE algorithms (\ref{MR_optimization_1}) and~(\ref{MR_optimization_2}) only address situations where the broken-down vehicles are positioned at the end of non-desired lanes, as shown in Fig.~\ref{emerging_1} later. For more general scenarios where vehicles may break down in arbitrary lanes, the C2TE algorithms cannot currently deal with this issue.  One potential approach involves the remaining vehicles stopping and maneuvering around the broken-down vehicles if sufficient space is available. However, such methods are highly case-specific and will require further investigation.
$\mathrm{(ii)}$ For another disturbance of the mixed-autonomy cases where some vehicles maintain a constant velocity, the C2TE algorithms (\ref{MR_optimization_1}) and (\ref{MR_optimization_2}) can partially handle it by excluding non-merging vehicles from the collision-avoidance CBF constraints in \eqref{stage_2_longitudinal_collision_avoidance_CBF}, i.e., 
${\partial \phi_{i,j}^{[2]}(x_i)}/{\partial x_i}(u_{i,x}^r-v_d)+\gamma\big(\phi_{i,j}^{[2]}(x_i)\big)-\delta_{i,j}^{[2]}\leq 0, j\in\mathcal N_i, j\notin \mathcal V_2,$ with $\mathcal V_2$ being the set of non-merging vehicles, as shown Fig.~\ref{different_lane_1}(c) later.
However, simply excluding non-merging vehicles may introduce potential collisions between vehicles, indicating further investigation as well. 
\end{remark}

\section{Experiments and extensive simulations}
In this section, we conduct real experiments and simulations to validate the effectiveness and flexibility of the C2TE algorithms (\ref{MR_optimization_1}) and (\ref{MR_optimization_2}), as shown in the online vedio \footnote{Online video: \href{https://www.youtube.com/watch?v=-M6kqAvOpOo}{https://www.youtube.com/watch?v=-M6kqAvOpOo}}.

\subsection{Experiments with three AMVs}
\label{sub_experiments}
\begin{figure}[!htb]
  \centering
  \includegraphics[width=6.3cm]{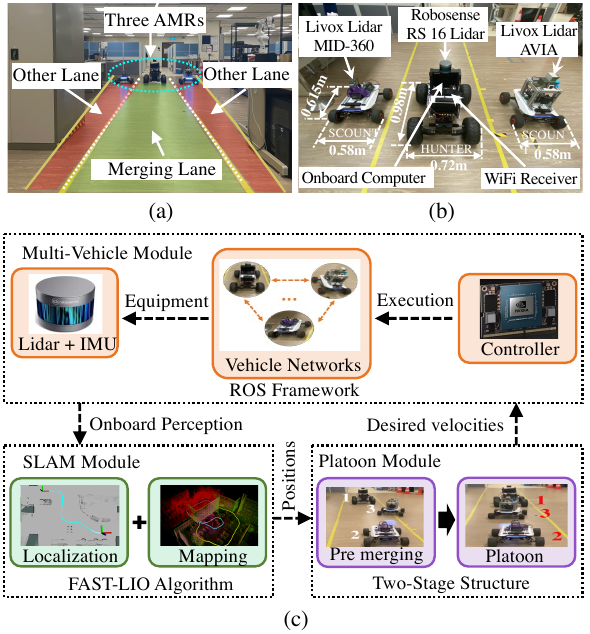}
  \caption{(a) The platoon merging working space. (b) Three AMVs and detailed components. (c) Structure of a multi-vehicle platoon merging system. }
  \label{Experiment_platform}
\end{figure}

We employ three autonomous mobile vehicles (AMVs) (i.e., one bigger HUNTER and two smaller SCOUTMINIs\footnote{HUNTER \& SCOUTMINI: \href{https://global.agilex.ai}{https://global.agilex.ai}}),
as shown in Fig.~\ref{Experiment_platform} (a)-(b). Each AMV is equipped with an onboard computer: NVIDIA Jetson Xavier NX3\footnote{NVIDIA Jetson Xavier NX3: \href{https://www.nvidia.com/en-us/autonomousmachines/
embedded-systems/jetson-xavier-nx/}{https://www.nvidia.com/en -us/autonomous machines/embedded-systems}}, an onboard Lidar: Robosense 16\footnote{Robosense RS 16: \href{https://www.robosense.ai/en/rslidar/RS-Helios}{https://www.robosense.ai/en/rslidar/}}, or Livox AVIA or Livox MID-360\footnote{Livox AVIA \& MID-360: \href{https://www.livoxtech.com}{https://www.livoxtech.com}}, and a WiFi Receiver. In Fig.~\ref{Experiment_platform} (c), we illustrate a multi-vehicle platoon merging structure. 
We first leverage the Lidar and IMU to conduct perception via a popular FAST-LIO SLAM algorithm~\cite{xu2022fast} to localize each AMV. After getting the positions, we send them to the platoon module to calculate desired velocities via a two-stage merging structure. Finally, the desired velocities $\{u_{i,x}^r, u_{i,y}^r\}$ are transferred to $\{v_i, \phi_i\}$ according to \eqref{velocity_convertion}.
We choose the safe and sensing distances $r=1.0$m, $R=1.1$m according to the vehicle base. The initial position and velocity of the virtual target in \eqref{virtual_target_dynamics} are set to be $[x_d(0), y_d(0)]\t=[3.5, 0.0]\t$m, and $v_d=0.2$m/s. The weight~$c$ in \eqref{MR_optimization_1} and \eqref{MR_optimization_2} is $c=100$.

\begin{figure*}[!htb]
  \centering
  \includegraphics[width=15.2cm]{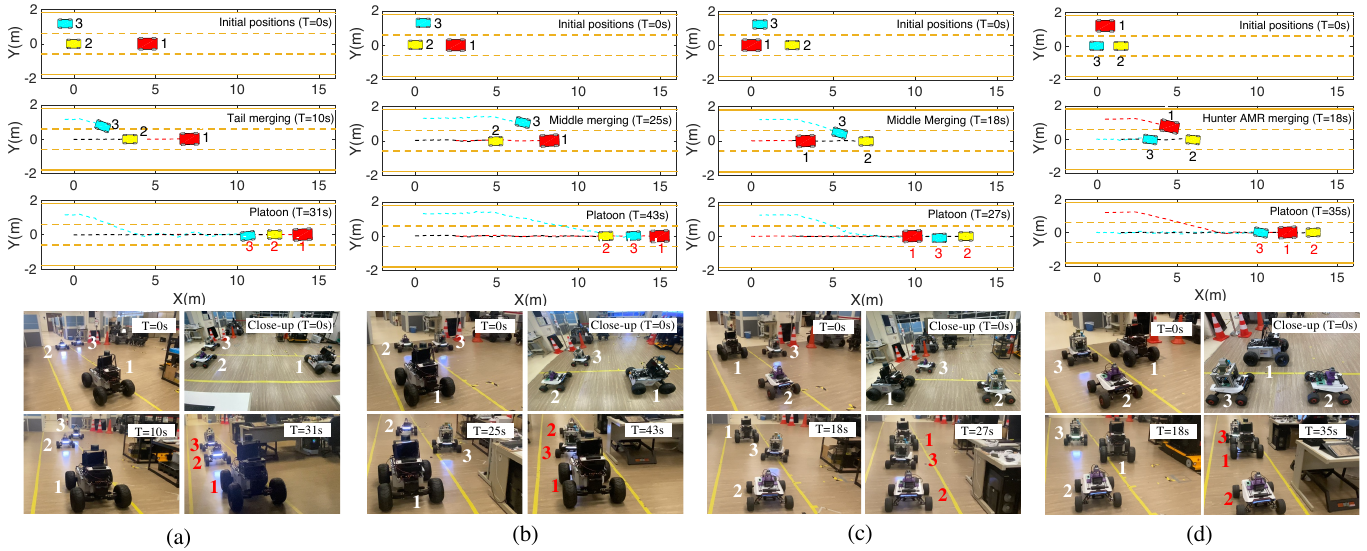}
  \caption{\textbf{First experiments: four cases of one-vehicle merging to show the flexibility of C2TE algorithms \eqref{MR_optimization_1} and \eqref{MR_optimization_2} to different initial positions.} (a)~Tail merging. (b) Middle merging 1. (c) Middle merging 2. (d) Middle merging 3. Here, the red, yellow and blue rectangles represent the HUNTER, SCOUTMINI 1, and SCOUTMINI 2, respectively, the dashed lines are the trajectories of the vehicles, the black and red numbers denote the initial and final positions, and the bottom four snapshots exhibit the one-vehicle merging experiments.  }
  \label{One_merging}
\end{figure*}

\begin{figure*}[!htb]
  \centering
  \includegraphics[width=15.2cm]{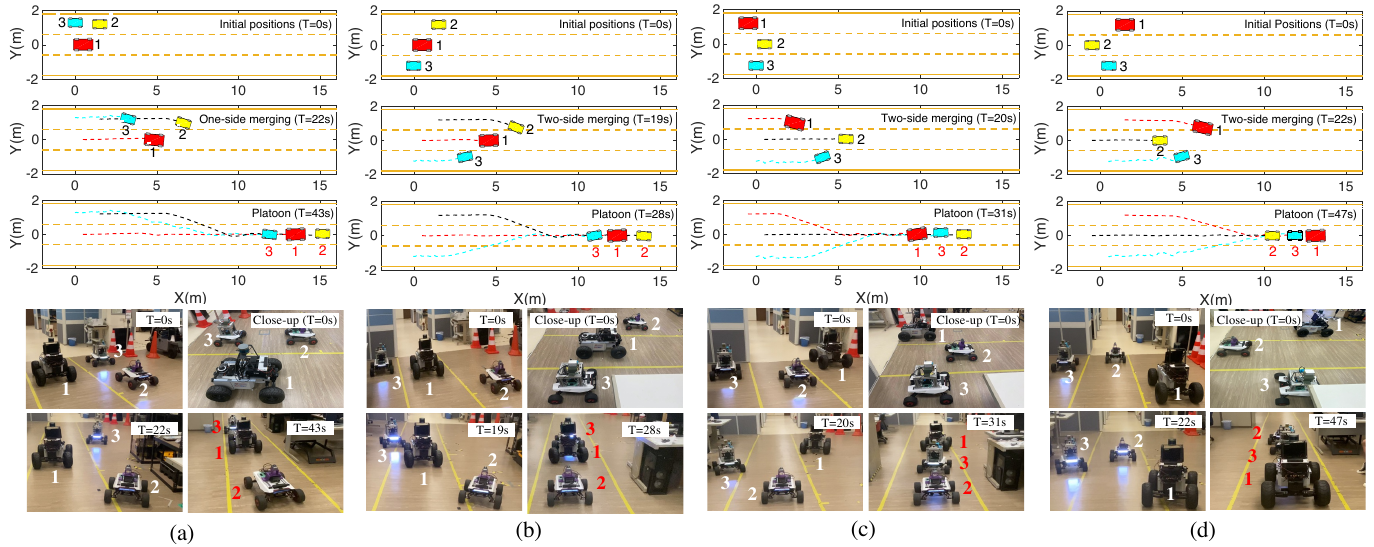}
  \caption{\textbf{Second experiments: four cases of two-vehicle merging to show the flexibility of C2TE algorithms \eqref{MR_optimization_1} and \eqref{MR_optimization_2} to different initial positions.} (a)~One-side merging. (b) Two-side merging 1. (c) Two-side merging 2. (d) Two-side merging 3. All the symbols have the same meanings as in Fig.~\ref{One_merging}.}
  \label{Two_merging}
\end{figure*}

\begin{figure}[!htb]
  \centering
  \includegraphics[width=6.2cm]{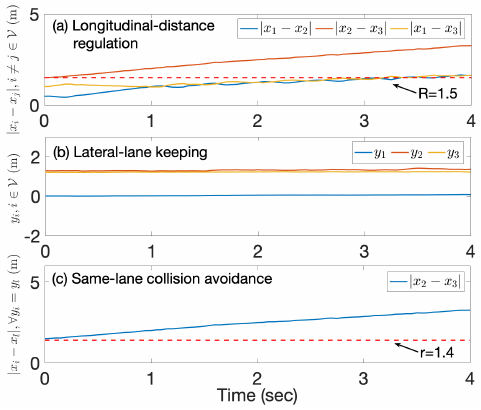}
  \caption{Temporal evolution of the states for pre-merging regulation in Definition~\ref{def_pre_merging}: The inter-vehicle longitudinal distances $|x_{i}(t)-x_j(t)|, i\neq j\in\mathcal V$, the lateral positions $y_i, i=1,2,3$, the same-lane inter-vehicle distance $|x_{2}-x_3|$ in Fig.~\ref{Two_merging} (a) for example.}
  \label{stage_1}
\end{figure}

\begin{figure}[!htb]
  \centering
  \includegraphics[width=6.2cm]{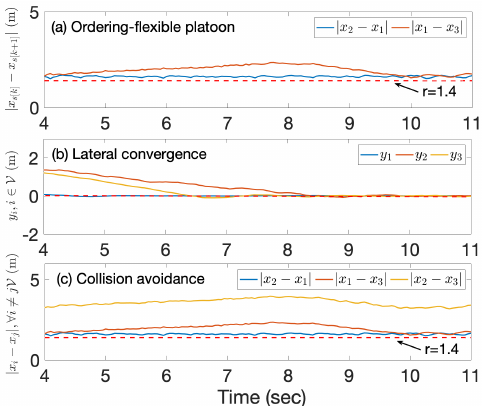}
  \caption{Temporal evolution of the states for {\it ordering-flexible} MVP merging in Definition~\ref{def_platoon}: The longitudinal errors of the platoon $|x_{s[k]}(t)-x_{s[k+1]}(t)|, i\in\mathbb{Z}_1^{N-1}$, the lateral position errors $y_i, i=1,2,3$, the inter-vehicle longitudinal distances $|x_{i}(t)-x_j(t)|, i\neq j\in\mathcal V$ in Fig.~\ref{Two_merging} (a) for example.}
  \label{stage_2}
\end{figure}

\begin{figure}[!htb]
  \centering
  \includegraphics[width=7.5cm]{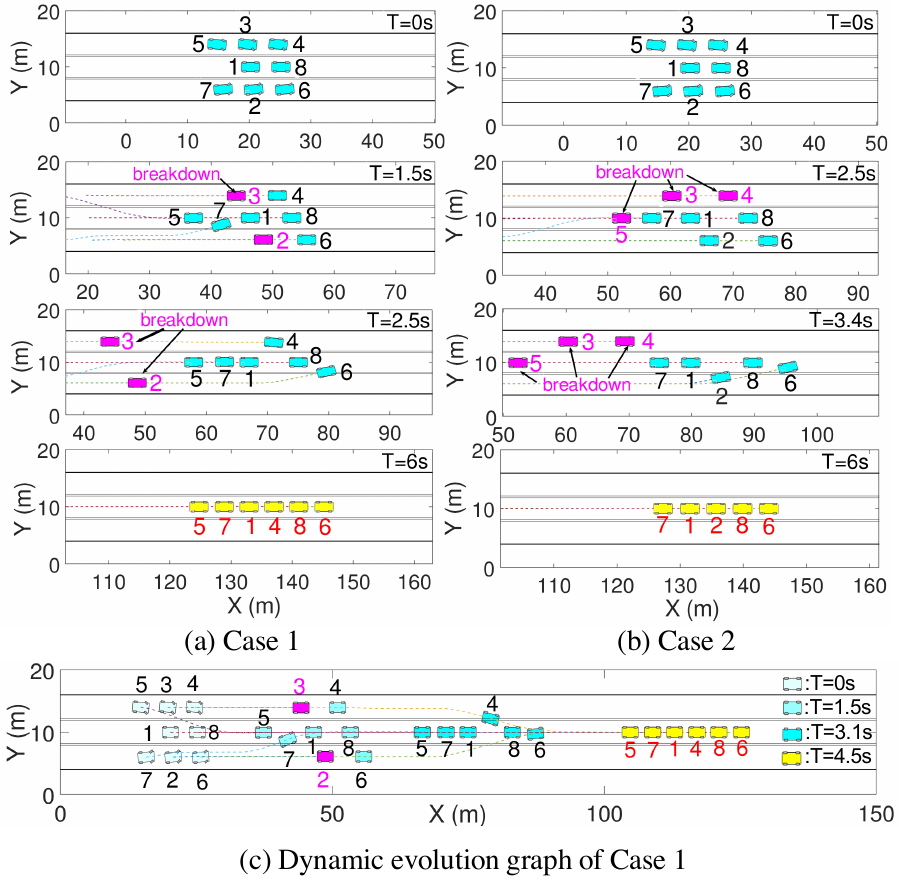}
  \caption{\textbf {First simulation: robustness to vehicles suddenly breaking down}. (a) Case 1: vehicles $i=2,3$ break down at $T = 1.5$s. (b) Case 2: vehicles $i=3,4,5$ break down at $T = 2.5$s. (c) Dynamic evolution graph of Case 1 to better show the vehicle merging performance. (Here, the blue and yellow rectangles denote the initial and final positions of normal vehicles, respectively, while the magenta rectangles denote the final positions of the broken vehicles. The dashed lines are the trajectories. The red numbers represent the distinct ordering sequences of the remaining vehicles.)}
  \label{break_1}
\end{figure}

\begin{figure}[!htb]
  \centering
  \includegraphics[width=7.5cm]{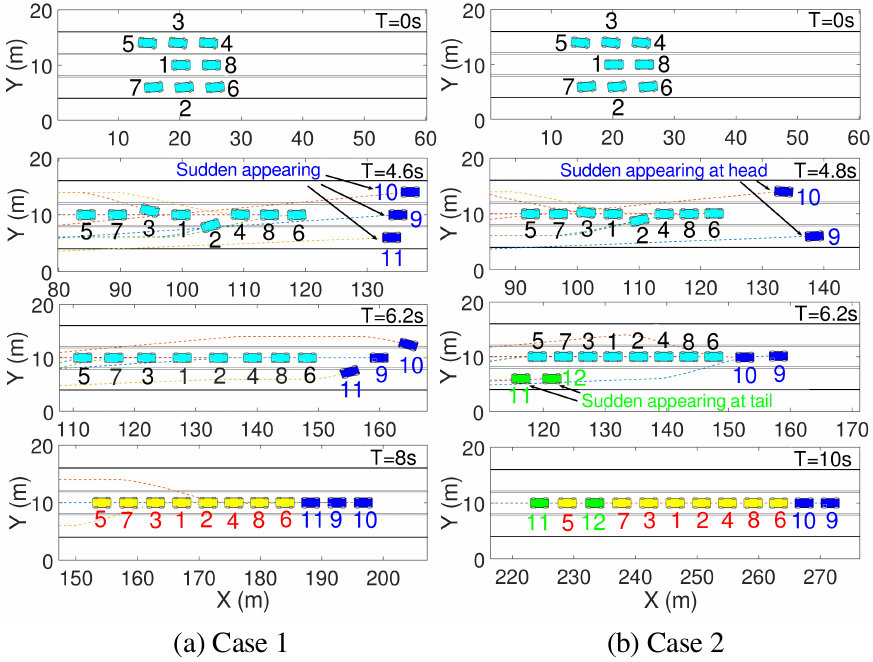}
  \caption{\textbf {Second simulation: adaptability to some new vehicles suddenly appearing}. (a) Case 1: vehicles $i=9,10,11$ appear in the front at $T = 4.6$s. (b) Case 2: vehicles $i=9,10$ appear in front of all the vehicles at $T = 4.8$s, and another two vehicles $i=11, 12$ appear on the behind of all the vehicles at $T=6.2$s. The blue, and yellow rectangles and dashed lines have the same meanings as in Fig. \ref{break_1}. The blue and green rectangles denote the emerging vehicles at the head, and tail, respectively. The blue and green numbers denote the different ordering compared with Fig. \ref{break_1}.  }
  \label{emerging_1}
\end{figure}

\begin{figure}[!htb]
  \centering
  \includegraphics[width=7.5cm]{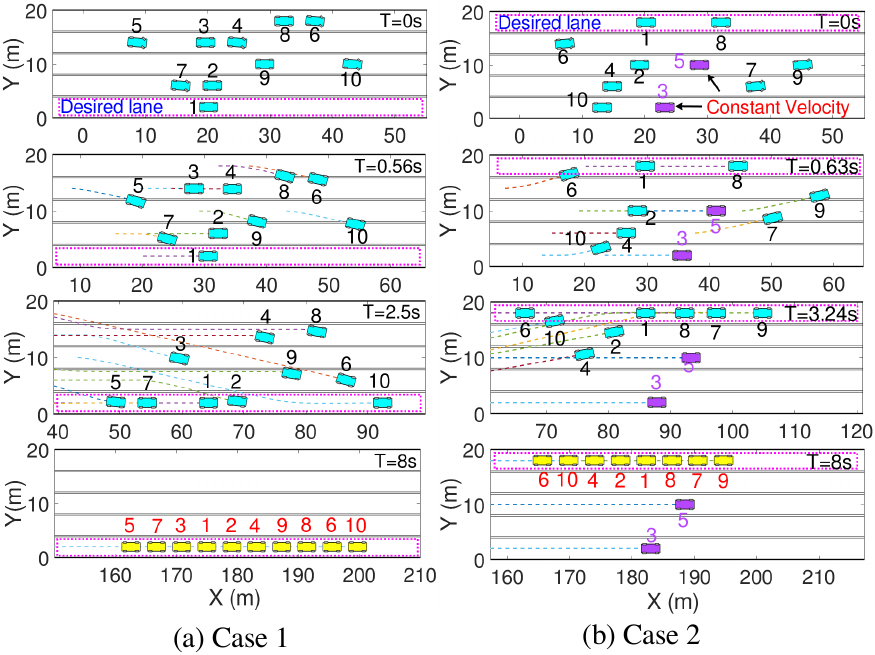}
  \caption{ \textbf {Third simulation: adaptability to a different number of lanes, choice of the desired lane, and mixed-autonomy cases}. (a) Case 1: The number of lanes is $5$, and the desired lane is placed at the bottom. (b) Case 2: The number of lanes is $5$, the desired lane is placed at the top, and the mixed vehicles $3,5$ maintain a constant velocity and do not execute merge missions. The blue and yellow rectangles, the black and red numbers, and the dashed lines have the same meanings as in Fig.~10 of the revised version. The purple rectangle represents mixed vehicles that maintain a constant velocity. The magenta dashed rectangle denotes the desired lane.} 
  \label{different_lane_1}
\end{figure}

\begin{figure}[!htb]
  \centering
  \includegraphics[width=\hsize]{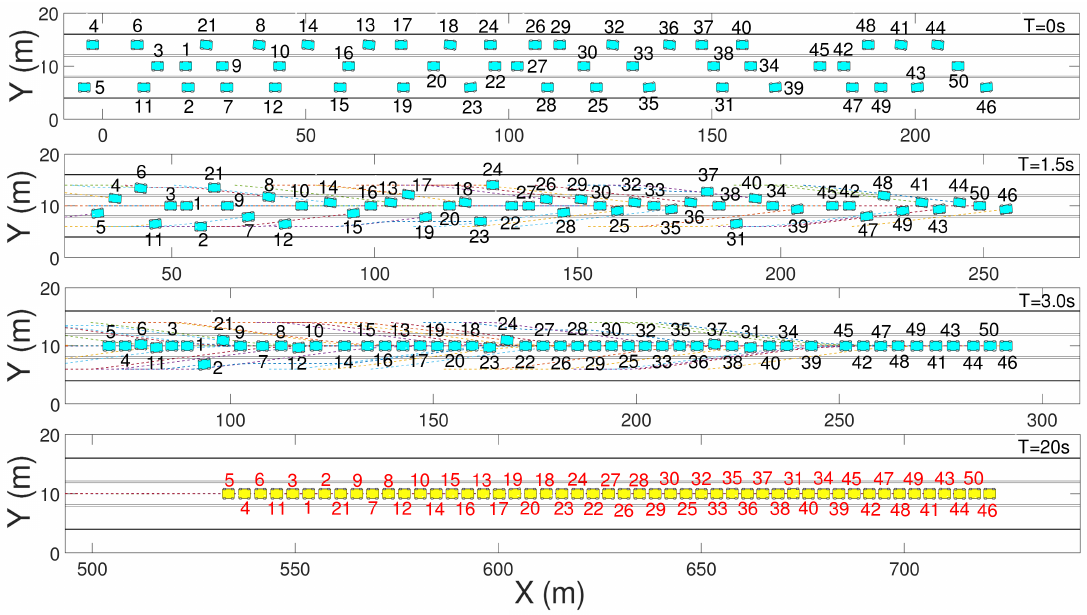}
  \caption{\textbf { Fourth simulation: scalability to the large-scale scenarios}. Trajectories of fifty vehicles from different initial lanes to the {\it ordering-flexible} MVP merging. The blue and yellow rectangles have the same meanings as in Fig.~\ref{break_1}.  }
  \label{vehicle_50}
\end{figure}


 For the one-vehicle merging experiments, we consider one-tail merging and three middle merging. Fig.~\ref{One_merging} (a)-(d) illustrates that three AMVs from different initial positions enlarge sufficient longitudinal distances and then merge into the platoon with distinct orderings (see Fig.~\ref{One_merging} (a): $\{1, 2, 3\}$, (b): $\{1, 3, 2\}$, (c): $\{2, 3, 1\}$ and (d): $\{2, 1, 3\}$).  For the two-vehicle merging experiments, we consider one-side merging and two-side mergings. Fig.~\ref{Two_merging} (a)-(d) also illustrates that three AMVs achieve the platoon (red numbers) with distinct orderings (see Fig.~\ref{Two_merging} (a): $\{2, 1, 3\}$, (b): $\{2, 1, 3\}$, (c): $\{2, 3, 1\}$ and (d): $\{1, 3, 2\}$). It is observed from Fig.~\ref{Two_merging} (b)-(d) that the flexibility to handle more complex situations can still be handled as well. 
Additionally, we take Fig.~\ref{Two_merging} (a) as an example to analyze the detailed evolution. It is observed in Fig.~\ref{stage_1}~(a) that $|x_i-x_j|\geq 1.5$ for $t>4$s, which ensures Objective~\ref{Pre_platoon_obj_1}) in Definition~\ref{def_pre_merging}. The lateral positions $y_i$ keeping invariant in Fig.~\ref{stage_1} (b) verifies Objective~\ref{Pre_platoon_obj_2}) of Definition~\ref{def_pre_merging}. The longitudinal distance between vehicles $2,3$ satisfies $|x_2(t)-x_3(t)|\geq1.4, \forall t>0$, which demonstrates Objective~\ref{Pre_platoon_obj_3}) of Definition~\ref{def_pre_merging}. Fig.~\ref{stage_2} (a) illustrates that $|x_{s[k]}(t)-x_{s[k+1]}(t)|=r, i\in\mathbb{Z}_1^{N-1}$ is satisfied in Objective~\ref{platoon_obj_2}) of Definition~\ref{def_platoon}. Fig.~\ref{stage_2} (b) illustrates that $\lim_{t\rightarrow\infty}y_i(t)=y_d, i=1,2,3$, which verifies Objective~\ref{platoon_obj_1}) of Definition~\ref{def_platoon}. It is observed in Fig.~\ref{stage_2}~(c) that the collision avoidance is guaranteed all along, i.e., Objective~\ref{platoon_obj_4}) of Definition~\ref{def_platoon}.



\subsection{Simulations}
\label{section_simulation}

We consider $N=8$ vehicles governed by~\eqref{vehicle_dynamics},   (\ref{MR_optimization_1}) and (\ref{MR_optimization_2}) with $B_i=2$m, $R=5$m, $r=3$m. The desired lane is $\mathcal L_i^d :=\{ \xi_i\in\mathbb{R}~|~\phi_i(\xi_i):=|\xi_i-10|=0 \}$, where the virtual target in \eqref{virtual_target_dynamics} is $[x_d(0), y_d(0)]\t=[20, 10]$m, and $v_d=20$m/s. The weight $c$ is the same as in Section~\ref{sub_experiments}.

To show the robustness to vehicles' breakdown, Fig.~\ref{break_1} illustrates two special cases of vehicles suddenly breakdown. As shown in Fig.~\ref{break_1} (a), the rest of six vehicles $i=1,4,5,6,7,8$ satisfying Assumptions~\ref{assump_yaw_angle}-\ref{assum_radius}  form a platoon (yellow rectangles) with distinct orderings when vehicles $i=2,3$ (magenta rectangles) break down at $T=2.5$s. Analogously,  Fig.~\ref{break_1}~(b) illustrates that the remaining five vehicles $i=1,2,6,7,8$ can achieve the platoon when vehicles $i=3,4,5$ break down at $T=2.5$s. To showcase the adaptability in tackling new emerging vehicles, we consider two specific cases in Fig.~\ref{emerging_1}. For Case 1 in Fig.~\ref{emerging_1} (a), eight vehicles initially form an {\it ordering-flexible} platoon. When three vehicles $i=9, 10, 11$ appear in the front at $T=4.6$s, the original eight-vehicle platoon will form the eleven-vehicle platoon. For Case 2 in Fig.~\ref{emerging_1} (b), we consider two new vehicles $i=9,10$ appearing with different positions, where vehicles $i=9,10$ are in different orderings compared with Fig.~\ref{emerging_1} (a). When another two vehicles $i=11,12$ suddenly emerge at the tail, the twelve-vehicle platoon will be achieved with different orderings. 
To demonstrate the adaptability for an arbitrary number of lanes and desired lane, Figs.~\ref{different_lane_1} (a)-(b) illustrate that ten vehicles can cross multiple lanes and form an {\it ordering-flexible} platoon (magenta dashed rectangle). Moreover, to show the two-phase transition, in Fig.~\ref{different_lane_1}(a) when $t = 0.56$ s, vehicles $5, 6, 7, 8, 9, 10$ have already entered Stage 1, while vehicles $2, 3, 4$ remain in Stage 1 due to insufficient longitudinal space. Moreover, for the mixed-autonomy cases in Fig.~\ref{different_lane_1}(b), the merging vehicles $1, 2, 4, 6, 7, 8, 9, 10$ form an {\it ordering-flexible} platoon, even when non-merging vehicles $3,5$ maintain a constant velocity.
Additionally, to show the scalability of large-scale scenarios, we conduct $n=50$ vehicles. Fig.~\ref{vehicle_50} illustrates that fifty vehicles form the {\it ordering-flexible} MVP on the desired lane, which demonstrates that the proposed C2TE algorithms (\ref{MR_optimization_1}) and (\ref{MR_optimization_2}) can be scaled up to large-scale scenarios.

\section{Conclusion}
In this paper, we have presented the C2TE algorithm such that multiple vehicles enlarge sufficient longitudinal distances and merge into a platoon with arbitrary orderings. The flexibility, effectiveness, robustness, adaptability, and scalability of the proposed algorithm have been showcased through experiments and simulations. Future work will investigate the rigorous analysis of the finite-time MVP merging, obstacle avoidance, mix-autonomy scenarios, and maximizing-road-throughput strategies.

\section{ Appendix}
\subsection{Appendix A}
\label{Appendix_A}
{\it Proof of Lemma~\ref{stage_2_lemma_lateral}.}
Let $\lambda_{i,c}^{[2]}\in\mathbb{R}^+$ be the Lagrange multiplier for the constraint (\ref{lateral_MR_optimization_2}b), we can design the corresponding Lagrangian function $\mathcal L_{i,c}^{[2]}$ for the optimization~\eqref{lateral_MR_optimization_2} to be $\mathcal L_{i,c}^{[2]}(u_{i,y}^r, \delta_{i,c}^{[2]}, \lambda_{i,c}^{[2]})=(u_{i,y}^r)^2+c(\delta_{i,c}^{[2]})^2+\lambda_{i,c}^{[2]}g_{i,c}^{[2]}$
with $g_{i,c}^{[2]}:=\frac{\partial \phi_{i,c}^{[2]}(y_i)}{\partial y_i}u_{i,y}^r+\gamma\big(\phi_{i,c}^{[2]}(y_i)\big)-\delta_{i,c}^{[2]}$. Since the optimization \eqref{lateral_MR_optimization_2} is convex, it follows from KKT condition \cite{boyd2004convex} that  $\frac{\partial \mathcal L_{i,c}^{[2]}(u_{i,y}^{r\ast}, \delta_{i,c}^{[2]\ast}, \lambda_{i,c}^{[2]\ast})}{\partial [u_{i,y}^r, \delta_{i,c}^{[2]}]\t}=\mathbf{0}_2, \lambda_{i,c}^{[2]\ast}g_{i,c}^{[2]\ast}=0, \lambda_{i,c}^{[2]\ast}\geq 0, g_{i,c}^{[2]\ast}\leq 0$
with $u_{i,y}^{r\ast}, \delta_{i,c}^{[2]\ast}, \lambda_{i,c}^{[2]\ast}$ being the optimal solutions. Then, one has that 
\begin{align}
\label{stage2_optimal_condition}
2u_{i,y}^{r\ast}+\lambda_{i,c}^{[2]\ast}\frac{\partial \phi_{i,c}^{[2]}(y_i)}{\partial y_i}=0, 2c\delta_{i,c}^{[2]\ast}-\lambda_{i,c}^{[2]\ast}=0.
\end{align}
We divide the lateral convergence into two cases.

\textbf{Case 1:} $\lambda_{i,c}^{[2]\ast}=0$.
If Case 1 holds, it follows from \eqref{stage2_optimal_condition} that $u_{i,y}^{r\ast}=0, \delta_{i,c}^{[2]\ast}=0$. 
Combining with the definition of $g_{i,c}^{[2]}$, one has that $\gamma\big(\phi_{i,c}^{[2]}(y_i)\big)\leq0$.
Since $\gamma(\cdot)$ is the extended class $\mathcal K$ function in \eqref{constraint_minimization} and $\phi_{i,c}^{[2]}(y_i)=|y_i-y_d|\geq0$ in \eqref{stage_2_longitudinal_merging_CBF}, one has that $\phi_{i,c}^{[2]}(y_i)=0$, i.e., 
$y_i=y_d$, which implies that vehicle $i$ is already on $\mathcal L_i^d$, i.e., $y_i=y_d$.

\textbf{Case 2:} $\lambda_{i,c}^{[2]\ast}>0$.
If Case 2 holds, it follows from $\lambda_{i,c}^{[2]\ast}g_{i,c}^{[2]}=0$  
that the boundary of the constraint (\ref{lateral_MR_optimization_2}) is activated, i.e., $g_{i,c}^{[2]}=0$. Meanwhile, 
it follows from \eqref{stage2_optimal_condition} that $u_{i,y}^{r\ast}=-c\frac{\partial \phi_{i,c}^{[2]}(y_i)}{\partial y_i}\delta_{i,c}^{[2]\ast}$.
Substituting $u_{i,y}^{\ast}$ into $g_{i,c}^{[2]}=0$ yields $\delta_{i,c}^{[2]\ast}=\gamma\big(\phi_{i,c}^{[2]}(y_i)\big)\Xi_{i,c}^{[2]}$
with $\Xi_{i,c}^{[2]}=(1+c(\frac{\partial \phi_{i,c}^{[2]}(y_i)}{\partial y_i})^2)^{-1}$, the optimal lateral inputs $u_{i,y}^{r\ast}$ become
 \begin{align}
\label{stage2_lateral_input1}
u_{i,y}^{r\ast}=&-c\frac{\partial \phi_{i,c}^{[2]}(y_i)}{\partial y_i}\gamma\big(\phi_{i,c}^{[2]}(y_i)\big)\Xi_{i,c}^{[2]}.
\end{align}
We pick a Lyapunov function candidate
\begin{align}
\label{stage_2_lateral_V}
V_1^{[2]}(t)=\frac{1}{2}\sum_{i\in\mathcal V}\gamma\big(\phi_{i,c}^{[2]}(y_i)\big)^2.
\end{align}
By selecting a simple function $\gamma(\phi)=\phi$, it follows from \eqref{decouple_dynamics}, \eqref{stage_2_longitudinal_merging_CBF}, \eqref{stage2_lateral_input1}, \eqref{stage_2_lateral_V} that the derivative of $V_1^{[2]}$ becomes $\dot{V}_1^{[2]}(t)=-c\sum_{i\in\mathcal V}(\phi_{i,c}^{[2]}(y_i)\frac{\partial \phi_{i,c}^{[2]}(y_i)}{\partial y_i})^2 \Xi_{i,c}^{[2]}$
with ${\partial \phi_{i,c}^{[2]}(y_i)}/{\partial (y_i-y_d)}={\partial \phi_{i,c}^{[2]}(y_i)}/{\partial y_i}$.
Recalling $\lambda_{i,c}^{[2]\ast}>0$ in Case~2, it follows from Eqs.~\eqref{proprtty1} and \eqref{stage_2_lateral_converge} that $y_i\notin \mathcal T_{i,c}^{[2]}$, i.e., $\phi_{i,c}^{[2]}(y_i)\neq0$. Together with $\Xi_{i,c}^{[2]}>0$ below \eqref{stage2_lateral_input1}, one has that $\dot{V}_1^{[2]}=0$ only if $\phi_{i,c}^{[2]}(y_i)=0$ (i.e., $y_i=y_d$), which implies that largest invariance set in $\{y_i|\dot{V}_1^{[2]}=0\}$ only contains $y_d$. Therefore, the invariance set is compact, which follows from LaSalle's invariance principle \cite{khalil2002nonlinear} that $\lim_{t\rightarrow\infty} y_i(t)=y_d$. By combining Cases 1-2 together, the proof of lateral convergence is completed.

\subsection{Appendix B}
\label{Appendix_B}
{\it Proof of Lemma~\ref{stage_2_lemma_longitudinal_coolision}}
Firstly, let $N_i$ be the cardinality of the set $\mathcal N_i$, we define $\bm\delta_{i,g}^{[2]}:=[\delta_{i,m}^{[2]}, \overbrace{\cdots, \delta_{i,j}^{[2]}, \cdots}^{j\in \mathcal N_i}]\t\in\mathbb{R}^{N_i+1}
\Phi_{i,g}^{[2]}:=[\phi_{i,m}^{[2]}(x_i), \cdots, \phi_{i,j}^{[2]}(x_i),  \cdots ]\t\in\mathbb{R}^{N_i+1}, $ $\gamma(\Phi_{i,g}^{[2]}):=[\gamma(\phi_{i,m}^{[2]}(x_i)), \cdots, \gamma(\phi_{i,j}^{[2]}(x_i)),  \cdots ]\t\in\mathbb{R}^{N_i+1},$
where $x_i$ is dropped for compactness of notation, the longitudinal optimization \eqref{longitudinal_MR_optimization_2} becomes
\begin{align}
\label{longitudinal_MR_optimization_3}
&\min\limits_{u_{i,x}^r,\bm\delta_{i,g}^{[2]}}\Big\{(u_{i,x}^r-v_d)^2+c(\bm\delta_{i,g}^{[2]})\t\bm\delta_{i,g}^{[2]}\Big\},\nonumber\\
\mathrm{s.t.}&~\mathbf{g}_{i,g}^{[2]}\leq\mathbf{0}_{N_i+1}
\end{align}		
with $\mathbf{g}_{i,g}^{[2]}:=[g_{i,1}^{[2]}, \cdots, g_{i,n}^{[2]}]\t=\frac{\partial \Phi_{i,g}^{[2]} }{\partial x_i} (u_{i,x}^r-v_d)+\gamma(\Phi_{i,g}^{[2]})+\bm\delta_{i,g}^{[2]}$. Let $\bm\lambda_{i,g}^{[2]}:=[\lambda_{i,1}^{[2]}, \lambda_{i,2}^{[2]}, \cdots,  \lambda_{i,N_i+1}^{[2]}]\in\mathbb{R}^{N_i+1}$ be the Lagrangian multiplier with each term satisfying $\lambda_{i,l}^{[2]}\geq 0, l\in\mathbb{Z}_1^{N_i+1}$, the corresponding Lagrangian function for \eqref{longitudinal_MR_optimization_3} to be $\mathcal L_{i,g}^{[2]}(u_{i,x}^r, \bm\delta_{i,g}^{[2]}, \bm\lambda_{i,g}^{[2]})=(u_{i,x}^r-v_d)^2+c(\bm\delta_{i,g}^{[2]})\t\bm\delta_{i,g}^{[2]}+(\bm\lambda_{i,g}^{[2]})\t\mathbf{g}_{i,g}^{[2]}$.
Analogous to Appendix~\ref{Appendix_A} of Lemma~\ref{stage_2_lemma_lateral}, since the optimization  \eqref{longitudinal_MR_optimization_3} is convex, it follows form KKT condition \cite{boyd2004convex} that 
\begin{align}
\label{KKT_long}
&\frac{\partial \mathcal L_{i,g}^{[2]}(u_{i,x}^{r\ast}, \bm\delta_{i,g}^{[2]\ast}, \bm\lambda_{i,g}^{[2]\ast})}{\partial [u_{i,x}^r, (\bm\delta_{i,g}^{[2]})\t]\t}=\mathbf{0}_{N_i+1},~(\bm\lambda_{i,g}^{[2]\ast})\t\mathbf{g}_{i,g}^{[2]}=0, \nonumber\\
&\mathbf{g}_{i,g}^{[2]}\leq \mathbf{0}_{N_i+1}, \bm\lambda_{i,g}^{[2]\ast}\geq \mathbf{0}_{N_i+1}
\end{align}
with $u_{i,x}^{r\ast}, \bm\delta_{i,g}^{[2]\ast}, \bm\lambda_{i,g}^{[2]\ast}$ being the optimal solutions of \eqref{longitudinal_MR_optimization_3}. 
From the stationary condition in \eqref{KKT_long}, one has that
\begin{align}
\label{stage2_long_input}
u_{i,x}^{r\ast}=-\frac{1}{2}\frac{\partial (\Phi_{i,g}^{[2]})\t}{\partial x_i} \bm\lambda_{i,g}^{[2]\ast}+v_d,~\bm\delta_{i,g}^{[2]\ast}=\frac{\bm\lambda_{i,g}^{[2]\ast}}{2c}.
\end{align}
Since $\lambda_{i,1}^{[2]\ast}, \lambda_{i,l}^{[2]\ast}, l\in\mathbb{Z}_2^{N_i+1}$ are associated with (\ref{longitudinal_MR_optimization_2}a) and (\ref{longitudinal_MR_optimization_2}b), we hereby divide the proof into four cases.

\textbf{Case 1:} $\bm\lambda_{i,g}^{[2]\ast}=\mathbf{0}_{N_i+1}$.
If Case 1 holds, the optimal solutions in \eqref{stage2_long_input} become $u_{i,x}^{r\ast}=v_d, \bm\delta_{i,g}^{[2]\ast}=\mathbf{0}_{N_i+1}$. Substituting $u_{i,x}^{r\ast}, \bm\delta_{i,g}^{[2]\ast}$ into the constraints $\mathbf{g}_{i,g}^{[2]}\leq \mathbf{0}_{N_i+1}$ in~\eqref{longitudinal_MR_optimization_3}, one has that $\gamma(\Phi_{i,g}^{[2]})\leq \mathbf{0}_{N_i+1}$, which implies that 
$\Phi_{i,g}^{[2]}\leq \mathbf{0}_{N_i+1}\Rightarrow x_i=x_d, r-|x_i-x_j|\leq 0$ because of the extended class $\mathcal K$ function $\gamma(\cdot)$, and \eqref{stage_2_longitudinal_merging_CBF}, \eqref{stage_2_longitudinal_collision_avoidance_err}. Therefore, the collision avoidance is guaranteed in Case 1.

\textbf{Case 2:} $\lambda_{i,1}^{[2]\ast}>0, \lambda_{i,l}^{[2]\ast}=0, \forall l\in\mathbb{Z}_2^{N_i+1}.$
If Case 2 holds, using the relation of $\lambda^{\ast}$ and $x(t) \in\mathcal T$ in Remark~\ref{remark_boundary}, one has that $\lambda_{i,l}^{[2]\ast}=0, \forall l\in\mathbb{Z}_2^{N_i+1} \Rightarrow x_i\in\mathcal T_{i,j}^{[2]}, \forall j\in\mathcal N_i$. According to the definition of $\mathcal T_{i,j}^{[2]}$, one has that $r-|x_i-x_j|\leq 0, \forall j\in\mathcal N_i$, which implies that the neighboring collision avoidance is thus guaranteed in Case 2 as well.

\textbf{Case 3:} $\lambda_{i,1}^{[2]\ast}=0,  \lambda_{i,l}^{[2]\ast}>0, \exists l\in\mathbb{Z}_2^{N_i+1}.$
If Case 3 holds, there exists $\lambda_{i,q}^{[2]\ast}=0, \forall q\neq l\in\mathbb{Z}_2^{N_i+1}$. Analogous to Case 2, it follows from \eqref{proprtty1} that $r-|x_i-x_q|\leq 0$, which implies that the collision avoidance between vehicles $i$ and $q$ is thus guaranteed.
For $\lambda_{i,l}^{[2]\ast}>0, \exists l\in\mathbb{Z}_2^{N_i+1}$, let $\mathcal A_i^{[2]}\notin \emptyset\subseteq \mathcal N_i$ for which  $\lambda_{i,l}^{[2]\ast}>0$ and $a_i=|\mathcal A_i^{[2]}|$ be the cardinality of the set $\mathcal A_i^{[2]}$, we can define $\widetilde{\bm\lambda}_{i,3}^{[2]}:=[ \cdots, \lambda_{i,l}^{[2]},$ $ \cdots]\t\in\mathbb{R}^{a_i}, \widetilde{\mathbf{g}}_{i,3}^{[2]}:=[\cdots, g_{i,l}^{[2]}, \cdots]\t\in\mathbb{R}^{a_i}, \widetilde{\bm\delta}_{i,3}^{[2]}:=[\cdots, \delta_{i,l}^{[2]},  \cdots]\t$ $\in\mathbb{R}^{a_i}, \widetilde{\Phi}_{i,3}^{[2]}:=[\cdots, \phi_{i,l}^{[2]}(x_i),  \cdots ]\t\in\mathbb{R}^{a_i}, \gamma(\widetilde{\Phi}_{i,3}^{[2]}):=[$ $\gamma(\phi_{i,m}^{[2]}(x_i)), \cdots, \gamma(\phi_{i,l}^{[2]}(x_i)),  \cdots ]\t\in\mathbb{R}^{a_i}$, the optimal solutions in~\eqref{stage2_long_input} degenerate to be $u_{i,x}^{r\ast}=-\frac{1}{2}\frac{\partial (\widetilde{\Phi}_{i,3}^{[2]})\t}{\partial x_i} \widetilde{\bm\lambda}_{i,3}^{[2]\ast}+v_d,~\widetilde{\bm\delta}_{i,3}^{[2]\ast}=\frac{\widetilde{\bm\lambda}_{i,3}^{[2]\ast}}{2c}.$
Using $(\widetilde{\bm\lambda}_{i,3}^{[2]\ast})\t\widetilde{\mathbf{g}}_{i,3}^{[2]}=0$ in \eqref{KKT_long}, one has that $\widetilde{\mathbf{g}}_{i,3}^{[2]}=\mathbf{0}_{a_i}$. Substituting $u_{i,x}^{r\ast}$ into $\widetilde{\mathbf{g}}_{i,3}^{[2]}=\mathbf{0}_{a_i}$ 
yields $\widetilde{\bm\delta}_{i,3}^{[2]\ast}=\Delta_{i,3}^{[2]}\gamma(\widetilde{\Phi}_{i,3}^{[2]})$ with $\Delta_{i,3}^{[2]}=(c\frac{\partial \widetilde{\Phi}_{i,3}^{[2]} }{\partial x_i}\frac{\partial (\widetilde{\Phi}_{i,3}^{[2]})\t}{\partial x_i} )+I_{a_i})^{-1}\in\mathbb{R}^{a_i\times a_i},$
which implies that 
\begin{align}
\label{stage2_long_input2_case3}
u_{i,x}^{r\ast}=-c\frac{\partial (\widetilde{\Phi}_{i,3}^{[2]})\t}{\partial x_i} \Delta_{i,3}^{[2]}\gamma(\widetilde{\Phi}_{i,3}^{[2]})+v_d.
\end{align}
We pick a Lyapunov function candidate 
\begin{align}
\label{stage2_V_case3}
V_2^{[2]}=\frac{1}{2}\sum_{i\in\mathcal V}\sum_{j\in\mathcal A_i}\gamma(\phi_{i,j}^{[2]}(x_i))^2.
\end{align}
By selecting a simple function $\gamma(\phi)=\phi$ and taking the derivative of $V_2^{[2]}$ along \eqref{decouple_dynamics} yields $\dot{V}_2^{[2]}=2\sum_{i\in\mathcal V}\sum_{j\in\mathcal A_i}\phi_{i,j}^{[2]}(x_i)\frac{\partial \phi_{i,j}^{[2]}(x_i)}{\partial x_i}u_{i,x}^r$
with ${\partial \phi_{i,j}^{[2]}(x_i)}/{\partial x_i}={\partial \phi_{i,j}^{[2]}(x_i)}/{\partial (x_i-x_j)}$. According to $\phi_{i,j}^{[2]}(x_i)$ in \eqref{stage_2_longitudinal_collision_avoidance_err} and  $\mathcal N_i$ in \eqref{neighbor_set}, one has that $\sum_{i\in\mathcal V}\sum_{j\in\mathcal A_i}\phi_{i,j}^{[2]}(x_i)\frac{\partial \phi_{i,j}^{[2]}(x_i)}{\partial x_i}v_d=0$, which follows from \eqref{stage2_long_input2_case3}
\begin{align}
\label{stage2_dV3_case3}
\dot{V}_2^{[2]} =&-2c\sum_{i\in\mathcal V}(\widetilde{\Phi}_{i,3}^{[2]})\t \frac{\partial \widetilde{\Phi}_{i,3}^{[2]} }{\partial x_i}\frac{\partial (\widetilde{\Phi}_{i,3}^{[2]})\t}{\partial x_i} \Delta_{i,3}^{[2]}\widetilde{\Phi}_{i,3}^{[2]},
\end{align}
because of $\gamma(\widetilde{\Phi}_{i,3}^{[2]})=\widetilde{\Phi}_{i,3}^{[2]}$. Meanwhile, using the Woodbury matrix identity \cite{grama2003accuracy}, the matrix $\Delta_{i,3}^{[2]}$ in \eqref{stage2_long_input2_case3}  becomes
$\Delta_{i,3}^{[2]}=I_{a_i}-c\frac{\partial \widetilde{\Phi}_{i,3}^{[2]} }{\partial x_i}\widetilde{\Delta}_{i,3}^{[2]}\frac{\partial (\widetilde{\Phi}_{i,3}^{[2]})\t}{\partial x_i}$
with $\widetilde{\Delta}_{i,3}^{[2]}=(1+c\frac{\partial (\widetilde{\Phi}_{i,3}^{[2]})\t }{\partial x_i}\frac{\partial \widetilde{\Phi}_{i,3}^{[2]}}{\partial x_i})^{-1}\in\mathbb{R^+}$,
which implies that $\frac{\partial (\widetilde{\Phi}_{i,3}^{[2]})\t}{\partial x_i} \Delta_{i,3}^{[2]}=\widetilde{\Delta}_{i,3}^{[2]}\frac{\partial (\widetilde{\Phi}_{i,3}^{[2]})\t}{\partial x_i}$.
Then, $\dot{V}_2^{[2]}$ in \eqref{stage2_dV3_case3} becomes $\dot{V}_2^{[2]} 
 \leq-2c\sum_{i\in\mathcal V}\widetilde{\Delta}_{i,3}^{[2]}\bigg\|(\widetilde{\Phi}_{i,3}^{[2]})\t \frac{\partial \widetilde{\Phi}_{i,3}^{[2]} }{\partial x_i}\bigg\|^2\leq 0$
with the constant $\widetilde{\Delta}_{i,3}^{[2]}>0$. 
From Assumption~\ref{assum_radius}, $V_2^{[2]}(0)$ in \eqref{stage2_V_case3} is bounded.
Combining with $\dot{V}_2^{[2]} \leq 0$, it further gives that $V_2^{[2]}$ is bounded all along for $t>0$ in Case~3. However, since $\phi_{i,j}^{[2]}(x_i)$ in \eqref{stage_2_longitudinal_collision_avoidance_err} satisfies $\phi_{i,j}^{[2]}(x_i)\rightarrow+\infty$ if $|x_i-x_j|\rightarrow r, \forall j\in\mathcal A_i$, one has that $V_2^{[2]}\rightarrow+\infty$ if $|x_i-x_j|\rightarrow r, \forall j\in\mathcal A_i$, which implies that the neighboring collision avoidance is thus guaranteed in Case 3.

\textbf{Case 4:} $\lambda_{i,1}^{[2]\ast}>0, \lambda_{i,l}^{[2]\ast}>0, \exists l\in\mathbb{Z}_2^{N_i+1}.$
If Case~4 holds, one has that there exists $\lambda_{i,q}^{[2]\ast}=0, \forall q\neq l\in\mathbb{Z}_2^{N_i+1}$. It implies that the collision avoidance between vehicles $i$ and $q$ is guaranteed, i.e., $|x_i(t)-x_q(t)|\geq r, \forall t>0$, which is similar to Case 3. Moreover, analogous to the vector definitions in Case~3, we can also define $\widetilde{\bm\lambda}_{i,4}^{[2]}:=[\lambda_{i,1} \cdots, \lambda_{i,l}^{[2]}, \cdots]\t\in\mathbb{R}^{a_i+1}, \widetilde{\mathbf{g}}_{i,4}^{[2]}:=[g_{i,1}^{[2]}, \cdots, g_{i,l}^{[2]}, \cdots]\t\in\mathbb{R}^{a_i+1}, \widetilde{\bm\delta}_{i,4}^{[2]}:=[\delta_{i,m}^{[2]}, \cdots, \delta_{i,l}^{[2]},  \cdots]\t \in\mathbb{R}^{a_i+1}, \widetilde{\Phi}_{i,4}^{[2]}:=[\phi_{i,m}^{[2]}(x_i), \cdots,$ $ \phi_{i,l}^{[2]}(x_i),  \cdots ]\t\in\mathbb{R}^{a_i+1},$ $ \gamma(\widetilde{\Phi}_{i,4}^{[2]}):= [\gamma(\phi_{i,m}^{[2]}(x_i)), \cdots, $ $\gamma(\phi_{i,l}^{[2]}(x_i)),  \cdots ]\t\in\mathbb{R}^{a_i+1}$, it follows from \eqref{stage2_long_input} that the optimal solutions in~Case 4 become
\begin{align}
\label{stage2_long_input_case4}
u_{i,x}^{r\ast}=-c\frac{\partial (\widetilde{\Phi}_{i,4}^{[2]})\t}{\partial x_i} \Delta_{i,4}^{[2]}\widetilde{\Phi}_{i,4}^{[2]}+v_d
\end{align}
with $\gamma(\widetilde{\Phi}_{i,4}^{[2]})=\widetilde{\Phi}_{i,4}^{[2]}$ and $\Delta_{i,4}^{[2]}=(c\frac{\partial \widetilde{\Phi}_{i,4}^{[2]} }{\partial x_i}\frac{\partial (\widetilde{\Phi}_{i,4}^{[2]})\t}{\partial x_i} )+I_{a_i+1})^{-1}\in\mathbb{R}^{(a_i+1)\times (a_i+1)}$.
Then, we pick a Lyapunov function candidate 
\begin{align}
\label{stage2_V_case4}
V_3^{[2]}=\frac{1}{2}\sum_{i\in\mathcal V}\Big\{\gamma(\phi_{i,m}^{[2]}(x_i))^2+\sum_{j\in\mathcal A_i}\gamma(\phi_{i,j}^{[2]}(x_i))^2\Big\}.
\end{align}
Analogously, taking the derivative of $V_3^{[2]}$ in \eqref{stage2_V_case4} and substituting $u_{i,x}^{r\ast}$ in \eqref{stage2_long_input_case4} yields $\dot{V}_3^{[2]}=-2c\sum_{i\in\mathcal V}(\widetilde{\Phi}_{i,4}^{[2]})\t$ $\frac{\partial \widetilde{\Phi}_{i,4}^{[2]} }{\partial x_i}\frac{\partial (\widetilde{\Phi}_{i,4}^{[2]})\t}{\partial x_i} \Delta_{i,4}^{[2]}\widetilde{\Phi}_{i,4}^{[2]}$.
From the fact that $\Delta_{i,4}^{[2]}=I_{a_i+1}-c\frac{\partial \widetilde{\Phi}_{i,4}^{[2]} }{\partial x_i}\widetilde{\Delta}_{i,4}^{[2]}\frac{\partial (\widetilde{\Phi}_{i,4}^{[2]})\t}{\partial x_i}$
with $\widetilde{\Delta}_{i,4}^{[2]}=(1+c\frac{\partial (\widetilde{\Phi}_{i,4}^{[2]})\t }{\partial x_i}\frac{\partial \widetilde{\Phi}_{i,4}^{[2]}}{\partial x_i})^{-1}\in\mathbb{R^+},$
one has that 
\begin{align}
\label{stage2_dV1_case4}
\dot{V}_3^{[2]}\leq&-2c\sum_{i\in\mathcal V}\widetilde{\Delta}_{i,4}^{[2]}\bigg\|(\widetilde{\Phi}_{i,4}^{[2]})\t \frac{\partial \widetilde{\Phi}_{i,4}^{[2]} }{\partial x_i}\bigg\|^2\leq 0,
\end{align}
where $\widetilde{\Delta}_{i,4}^{[2]}$ is a positive constant as well. Since $V_3^{[2]}$  in \eqref{stage2_V_case4} also contains the collision avoidance errors $\phi_{i,j}(x_i)$ and $\dot{V}_3^{[2]}\leq0$, the remaining proof 
is similar to the one in Case 3, which is thus completed.
By combining Cases 1-4 together, the proof is thus completed. 
\eproof

\subsection{Appendix C}
\label{Appendix_C}
{\it Proof of Lemma~\ref{stage_2_lemma_longitudinal_crusing}.}
To align with the proof in Lemma~\ref{stage_2_lemma_longitudinal_coolision}, we also consider four cases.

\textbf{Case 1:} $\bm\lambda_{i,g}^{[2]\ast}=\mathbf{0}_{N_i+1}$.
If Case 1 holds, it follows from Case~1 of Lemma~\ref{stage_2_lemma_longitudinal_coolision} that the collision avoidance between arbitrary two vehicles $i$ and $j$ is guaranteed, i.e., $|x_i(t)-x_j(t)|\geq r, \forall i\neq j\in\mathcal V,  t\geq0$. Meanwhile, since $\lambda_{i,1}^{[2]}=0$ in Case 1, it follows from \eqref{proprtty1} that 
that $x_i\in \mathcal T_{i,m}^{[2]}, \forall i\in\mathcal V$, i.e., $x_i=x_d, \forall i\in\mathcal V$, which further implies that  each vehicle $i$ coincides with other vehicle. However, such a situation inevitably contradicts the collision avoidance of $|x_i(t)-x_j(t)|\geq r, \forall i\neq j\in\mathcal V,  t\geq0$, one has that Case~1 can be excluded.

\textbf{Case 2:} $\lambda_{i,1}^{[2]\ast}>0, \lambda_{i,l}^{[2]\ast}=0, \forall l\in\mathbb{Z}_2^{N_i+1}.$
If Case 2 holds, the optimal solutions in \eqref{stage2_long_input} will become $u_{i,x}^{r\ast}=-\frac{1}{2}\frac{\partial (\phi_{i,m}^{[2]})\t}{\partial x_i} \lambda_{i,1}^{[2]\ast}+v_d,~\delta_{i,m}^{[2]\ast}=\frac{\lambda_{i,1}^{[2]\ast}}{2c}$
with $\phi_{i,m}^{[2]}, \lambda_{i,1}^{[2]\ast}, \delta_{i,m}^{[2]\ast}$ given in \eqref{longitudinal_MR_optimization_3}. Meanwhile, it follows from $(\bm\lambda_{i,g}^{[2]\ast})\t\mathbf{g}_{i,g}^{[2]}=0$ in \eqref{KKT_long} that the boundary of $g_{i,1}^{[2]}$ in \eqref{longitudinal_MR_optimization_3} is activated, i.e., $\frac{\partial \phi_{i,m}^{[2]}(x_i)}{\partial x_i}(u_{i,x}^r-v_d)+\phi_{i,m}^{[2]}(x_i)-\delta_{i,m}^{[2]}= 0$
with $\gamma\big(\phi_{i,m}^{[2]}(x_i)\big)=\phi_{i,m}^{[2]}(x_i)$, which implies that $\delta_{i,m}^{[2]\ast}=\phi_{i,m}^{[2]}(x_i)\zeta_i^{[2]}$ with 
$\zeta_i^{[2]}=\big(({\partial \phi_{i,m}^{[2]}(x_i)}/{\partial x_i})^2+1)\big)^{-1}$, which implies that $u_{i,x}^{r\ast}=-c\frac{\partial (\phi_{i,m}^{[2]})\t}{\partial x_i} \phi_{i,m}^{[2]}(x_i)\zeta_i^{[2]}+v_d.$
It is observed that $u_{i,x}^{r\ast}$ will govern vehicle $i$ converge to $p_d$, which implies $x_i=x_j=x_d, \forall i\neq j\in\mathcal V$. Then, Case 2 is degenerated to Case 1 and thus can be excluded as well.


\textbf{Case 3:} $\lambda_{i,1}^{[2]\ast}=0,  \lambda_{i,l}^{[2]\ast}>0, \exists l\in\mathbb{Z}_2^{N_i+1}.$
If Case 3 holds, from $\lambda_{i,1}^{[2]\ast}=0$ and \eqref{proprtty1}, one has that $x_i\in \mathcal T_{i,m}^{[2]}, \forall i\in\mathcal V$, i.e., $x_i=x_d, \forall i\in\mathcal V$, which implies that vehicle $i$ already coincides with the virtual target. However, such a situation can not be maintained all along. From $\lambda_{i,l}^{[2]\ast}>0, \exists l\in\mathbb{Z}_2^{N_i+1}$, there always exists vehicles $k$ and $l$ satisfying the collision avoidance for $\lambda_{i,l}^{[2]\ast}=0$. Therefore, Case 3 is excluded as well.

\textbf{Case 4:} $\lambda_{i,1}^{[2]\ast}>0, \lambda_{i,l}^{[2]\ast}>0, \exists l\in\mathbb{Z}_2^{N_i+1}.$
 If Case 4 holds, it follows from \eqref{stage2_dV1_case4} that $\dot{V}_2^{[2]}\leq0$, which implies that $\dot{V}_2^{[2]}=0$ only if $(\widetilde{\Phi}_{i,4}^{[2]})\t ({\partial \widetilde{\Phi}_{i,4}^{[2]} }/{\partial x_i})=0$ because 
$\Delta_{i,4}^{[2]}$ is a positive constant. Analogous to the proof of Lemma~\ref{stage_2_lemma_lateral}, it follows from LaSalle's invariance principle \cite{khalil2002nonlinear} that
\begin{align}
\label{Lemma5_case_4_convergence}
\lim_{t\rightarrow\infty}(\widetilde{\Phi}_{i,4}^{[2]}(t))\t \frac{\partial \widetilde{\Phi}_{i,4}^{[2]} (t)}{\partial x_i (t)}=0.
\end{align}
Substituting \eqref{Lemma5_case_4_convergence} into \eqref{stage2_long_input_case4} yields $u_{i,x}^{r\ast}=v_d
, \forall i\in\mathcal V$, which completes the proof of $\lim_{t\rightarrow\infty}\dot{x}_i(t)=\lim_{t\rightarrow\infty}\dot{x}_j(t)=v_d, \forall i\neq j\in\mathcal V$.
By combining Cases 1-4 together, the platoon cruising is achieved.
\eproof

\subsection{Appendix D}
\label{Appendix_D}
{\it Proof of Lemma~\ref{stage_2_lemma_longitudinal_ordering}.} From $\lim_{t\rightarrow\infty}\dot{x}_i(t)=\lim_{t\rightarrow\infty}\dot{x}_j(t)=v_d, \forall i\neq j\in\mathcal V$ in  Lemma~\ref{stage_2_lemma_longitudinal_crusing}, the limiting value of the longitudinal positions between adjacent vehicles $i\neq j\in\mathcal V$ remain invariant, which results in a platoon $x_{s[1]}>x_{s[2]}>\cdots> x_{s[N]}$ with an arbitrary ordering $s[1], \cdots, s[N]$ from the head to the tail of the platoon.

For the left-side inequality of $r<\lim_{t\rightarrow\infty}|x_{s[k]}(t)-x_{s[k+1]}(t)|$, it follows from Lemma~\ref{stage_2_lemma_longitudinal_coolision} that $|x_i(t)-x_j(t)|\geq r, \forall i\neq j\in\mathcal V,  t\geq0$, which gives $\lim_{t\rightarrow\infty}|x_{s[k]}(t)-x_{s[k+1]}(t)|>0$ for arbitrary adjacent vehicles $s[k]$ and $s[k+1]$. 
Regarding the right-side inequality of $\lim_{t\rightarrow\infty}$ $|x_{s[k]}(t)-x_{s[k+1]}(t)|<\rho, \forall k\in\mathbb{Z}_1^{N-1}$, we prove by contradiction. We assume that there exists at least one pair of vehicles satisfying $|x_{s[q]}(t)-x_{s[q+1]}(t)|\geq\rho$, and analyze the situations in three claims.

{\it Claim i)}: $x_{s[q]}(t)>x_{s[q+1]}>x_d$. For vehicle $s[q]$, one has that $(\phi_{q,m}^{[2]}(t))\t \frac{\partial \phi_{q,m}^{[2]}(t)}{\partial x_q (t)}=x_q-x_d>0$. Meanwhile, since $|x_{s[q]}-x_{s[q+1]}|\geq\rho$, one has that $|x_{s[w]}-x_{s[q+1]}|\geq\rho$ for $w=1, \cdots, q-1$, which implies that the neighboring vehicle $s[e]$ satisfies $x_{s[q]}-x_{s[e]}<\rho$. From $\phi_{i,j}^{[2]}$ in \eqref{stage_2_longitudinal_collision_avoidance_err}, one has that $\sum_{l\in\mathcal A_q}(\phi_{q,l}^{[2]})\t \frac{\partial \phi_{q,l}^{[2]}}{\partial x_q}>0$, which implies that $(\phi_{q,m}^{[2]})\t \frac{\partial \phi_{q,m}^{[2]}}{\partial x_q }+\sum_{l\in\mathcal A_q}(\phi_{q,l}^{[2]})\t \frac{\partial \phi_{q,l}^{[2]}}{\partial x_q}>0$.
However, such a condition contradicts \eqref{Lemma5_case_4_convergence}. Claim i) is excluded.

{\it Claim ii)}: $x_{s[q]}> x_d\geq x_{s[q+1]}$. For vehicle $s[q]$, one has that $(\phi_{q,m}^{[2]})\t \frac{\partial \phi_{q,m}^{[2]}}{\partial x_q}=x_q-x_d>0$. Analogous to Claim~i), there exists $|x_{s[w]}-x_{s[q+1]}|\geq\rho$ for $w=1, \cdots, q-1$. Therefore, we can also exclude this situation as well.

{\it Claim iii)}: $x_d\geq x_{s[q]}> x_{s[q+1]}$. For vehicle $s[q+1]$, one has that $(\phi_{q+1,m}^{[2]})\t \frac{\partial \phi_{q+1,m}^{[2]}}{\partial x_{q+1} }=x_{q+1}-x_d<0$. Meanwhile, since $|x_{s[q]}-x_{s[q+1]}|\geq\rho$, one has that $|x_{s[w]}-x_{s[q]}|\geq\rho$ for $w=q+2, \cdots, N$, which implies that there only exists the neighboring vehicles $s[e]$ satisfying $x_{s[q]}-x_{s[e]}<\rho$. Then, it from \eqref{stage_2_longitudinal_collision_avoidance_err} that $\sum_{l\in\mathcal A_{q+1}}(\phi_{q+1,l}^{[2]})\t \frac{\partial \phi_{q+1,l}^{[2]}}{\partial x_{q+1}}<0$, which implies that $(\phi_{q+1,m}^{[2]})\t \frac{\partial \phi_{q+1,m}^{[2]}}{\partial x_{q+1}}+\sum_{l\in\mathcal A_{q+1}}(\phi_{q+1,l}^{[2]})\t \frac{\partial \phi_{q+1,l}^{[2]}}{\partial x_{q+1}}<0$.
It contradicts \eqref{Lemma5_case_4_convergence}. Therefore, Claim iii) is excluded as well. 
By combining Claims i)-iii) together, the right-side inequality of $\lim_{t\rightarrow\infty}$ $|x_{s[k]}(t)-x_{s[k+1]}(t)|<\rho, \forall k\in\mathbb{Z}_1^{N-1}$ is guaranteed all along. The proof is thus completed.
\eproof

\bibliographystyle{IEEEtran}
\bibliography{IEEEabrv,ref}

\end{document}